\documentclass[letterpaper,11pt]{article}

\usepackage{fancyhdr}

\usepackage[utf8]{inputenc}
\usepackage[T1]{fontenc}

\usepackage[english]{babel}
\usepackage{authblk}
\usepackage[margin=1in]{geometry}
\usepackage{amsfonts}       % blackboard math symbols
\usepackage{nicefrac}       % compact symbols for 1/2, etc.
\usepackage{microtype}      % microtypography
\usepackage{amsmath,bbm,bm,mathtools} 
\usepackage{float}
\usepackage{amssymb}
\usepackage{amsthm}
\usepackage{algorithm}
\usepackage{algorithmic}
\usepackage{booktabs}       % professional-quality tables
\usepackage{graphicx}
\usepackage{multirow}
\usepackage{subcaption}
\usepackage[colorlinks=true, allcolors=black]{hyperref}
\usepackage{xcolor}
\usepackage{comment}
\usepackage[shortlabels]{enumitem}
\usepackage[normalem]{ulem}

\usepackage{natbib}
\setcitestyle{authoryear,open={(},close={)}}
\theoremstyle{plain}

\def\eqref#1{equation~\ref{#1}}

\def\1{\bm{1}}

\DeclareMathAlphabet{\mathsfit}{\encodingdefault}{\sfdefault}{m}{sl}
\SetMathAlphabet{\mathsfit}{bold}{\encodingdefault}{\sfdefault}{bx}{n}

\definecolor{DarkGreen}{rgb}{0.1,0.5,0.1}
\definecolor{DarkRed}{rgb}{0.5,0.1,0.1}
\definecolor{DarkBlue}{rgb}{0.1,0.1,0.5}
\definecolor{Gray}{rgb}{0.2,0.2,0.2}

\makeatletter
\newcommand\footnoteref[1]{\protected@xdef\@thefnmark{\ref{#1}}\@footnotemark}
\makeatother

\title{\Large{Artificial intelligence surrogates for treatment effect estimation with before-and-after data}} 

\author{\large{Frances Dean$^{\dagger\S}$ ~~~ Anna Neufeld$^{*}$ ~~~ Joshua Barrios$^{\S}$ ~~~  Geoffrey H Tison$^{\S}$ ~~~ Ahmed Alaa$^{\dagger\S}$}}

\affil{
$^\dagger$University of California, Berkeley\\
$^\S$University of California, San Francisco \\
$^*$Williams College \\}

\usepackage{hyperref}
\hypersetup{
  colorlinks=true,
  linkcolor=blue,
  citecolor=blue,
  urlcolor=blue
}

\usepackage{url}

\usepackage{multirow}
\usepackage{makecell}
\usepackage{mdframed}

\usepackage{comment}
\usepackage[utf8]{inputenc} % allow utf-8 input
\usepackage[T1]{fontenc}    % use 8-bit T1 fonts
\usepackage{hyperref}       % hyperlinks
\usepackage{url}            % simple URL typesetting
\usepackage{booktabs}       % professional-quality tables
\usepackage{amsfonts}       % blackboard math symbols
\usepackage{nicefrac}       % compact symbols for 1/2, etc.
\usepackage{microtype}      % microtypography
\usepackage{xcolor}         % colors
\usepackage{tikz}
\usepackage{tikz-cd}
\usetikzlibrary{positioning}
\usepackage{soul}
\usepackage{amssymb}
\usepackage{amsthm}
\usepackage{wrapfig}
\usepackage{amsmath}

\theoremstyle{plain} 

\date{}

\begin{document}

\maketitle

\begin{abstract}
Estimating the causal effects of medical treatments is difficult when clinically important outcomes are costly to measure~or~require~long follow-up. Short-term or inexpensive \textbf{\textit{surrogate}}~outcomes~offer~a~potential alternative, but surrogate biomarkers~may be unavailable or difficult to identify. Advances in artificial intelligence (AI) have enabled increasingly accurate prediction of clinical outcomes from inexpensive, high-dimensional measurements, which creates an opportunity to use AI predictions themselves as surrogates.~To~this end, we develop a framework for estimating treatment~effects~from paired measurements obtained \textbf{\textit{before and after}} treatment~for~each treated individual. A pretrained AI~model~is~applied~to~the~before and after measurements, and our estimator compares the resulting outcome predictions. We characterize~the~ technical~assumptions~under~which this within-person contrast identifies the average treatment effect on the treated, even when clinical outcomes are never observed for treated individuals. When these assumptions~cannot~be~justified, we use prediction-powered inference to correct bias using a small number of observed clinical outcomes and obtain valid inference. Synthetic and real-world cardio-oncology experiments demonstrate the validity and accuracy of the approach.
\end{abstract}

\section{Introduction}

Clinical endpoints, such as mortality and advanced disease progression, are often costly, invasive, or long term. This slows medical innovation and pharmacovigilance. Researchers often use earlier, more readily accessible measurements as proxies for later outcomes. These are known as surrogate outcomes. AI models are increasingly trained on cheap, non-invasive, and short-term high dimensional data to predict clinical outcomes. Suggestions that we might use predictions from an AI model as a surrogate outcome are growing in attention \citep{baicker2026ai, kadakia2026adoption, woodford2026enhancing, rudroff2025digital, kruizinga2020development}. Suppose, for example, we want to monitor long term structural heart failure. In the short term, we do not have access to patient mortality or severe disease progression. Patients seek care, and we desire to intervene prior to irreversible disease, cheaply and rapidly knowing if our intervention worked for this individual. We might also hope to have a rapid clinical trial in the case of the development of new medicines. AI can now predict structural heart disease from electrocardiography (ECG), picking up signals suggestive of long-term disease \citep{poterucha2025detecting, hughes2026echonext}. Using a trained AI model shown to be predictive of heart failure from non-invasive, low-cost ECGs could allow for the assessment of changes in mortality risk in the short-term by comparing ECGs before and after treatment. This would be revolutionary for treatment planning and modification. {However, our ability to do this accurately relies not only on the AI model having high predictive accuracy for heart failure but also that the treatment under investigation changes heart failure risk through a mechanism captured by the AI model. This relationship is not obvious and potentially absent for many AI models; for instance, ECGs are not traditionally used to diagnose heart failure and the mechanisms by which ECG-AI detects heart failure risk may not be clear.} 

Traditional surrogate outcomes have a notorious history of failures \citep{fleming1996surrogate}. Bisphosphonates increase bone mineral density but increase atypical fracture risk \citep{black2020atypical}. Flecainide and encainide decreased minor arrhythmias but increased mortality and are now warned against or retracted \citep{echt1991mortality}. Many FDA accelerated approvals for drug candidates using surrogate endpoints do not reach full approval after trial with real clinical endpoints \citep{kim2015cancer, liu2024clinical}. Without careful evaluation and delineation of assumptions in using AI based surrogates, widespread adoption of this practice risks being highly problematic as the properties that make a variable a surrogate for statistical and causal analyses differ from what makes a good AI model with high performance on clinical prediction tasks.  

Studying the validity of AI-based surrogates for causal inference is of timely importance. Indeed, AI predictions are already being used as surrogates~to~study~causal effects \citep{dandelion_white_paper, sweeney2026development, pulaski2025clinical}. An AI-based pathology endpoint~for~Metabolic dysfunction-Associated Steatohepatitis (AIM-MASH) has even been FDA approved for use in trials \citep{pulaski2025clinical, mash_usa}.~Pathology based endpoints, like AIM-MASH, will be the start of a large number of AI surrogates utilized to expedite and automate outcome collection for causal analyses. 

Our paper proposes a framework for treatment effect estimation with AI surrogates using patients as their own controls. In our setting, treated patients have access to a high-dimensional measurement taken at two time points: once before and once after treatment. An AI model is trained on a separate dataset of untreated patients to learn the relationship between these measurements and the clinical endpoint. This model is then used to evaluate the predicted score for the clinical outcome~with~and~without treatment in the treated dataset using the before and after measurements, and the AI predictions serve as surrogate outcomes. In the example of heart failure, our framework is the following: we, or collaborators elsewhere,~have~a retrospective training dataset of ECGs for untreated patients with enough follow up data to have a large number of true heart failure mortality endpoints.~An~AI model is trained on this dataset. {For a treatment we seek to evaluate, we obtain an ECG from shortly before and sometime after the treatment is performed in treated patients. Each ECG is input into the model and the scores compared.} Our paper answers {if and when} this framework can be used to identify the causal effect of the treatment on heart failure mortality. We use ECGs as an example throughout our work, but the framework is broadly applicable.

\section{Problem Setup}

\subsection{Mathematical Notation}

Throughout, random variables are denoted with capital letters and their realizations with lower case letters. Superscripts indicate time indices and subscripts indicate individuals. We consider the setting where $T$ is time and a binary treatment $A^t \in \{0, 1\}$ is applied once so that $A^t=1$ after treatment and $A^t=0$ before. We observe a high-dimensional covariate matrix or measurement $X^t \in \mathcal{X}$, i.e. an ECG, which is affected by $A^t$ and predicts an outcome $Y^t$. The outcome of interest in our set up, $Y^t \in \mathbb{R}$, is {\it costly} to observe, i.e. heart failure is observed only by echocardiography or by death, hence AI predicts the outcome $Y^t$ from the measurement $X^t$. Subjects for inference on treatment effects are observed at two time points, $t \in \{t^{pre}_i, t^{post}_i\}$, between the two measurement times $A^t$ occurs, i.e. $A^{t^{post}}=1$ and $A^{t^{pre}}=0$. We observe $X^t$ at both time points, so that we have $X^t$ for each $t \in \{t^{pre}, t^{post}\}$. For each subject $i$, the AI will observe $X^{t^{pre}}_i, X^{t^{post}}_i$. The calendar times $t^{pre}, t^{post}$ can vary by individual, but the interval $t^{post}-t^{pre}$ is considered fixed. We do not have labels $Y^t$ for treated individuals due to cost. We call this unlabeled observational dataset of some $n$ patients $\mathcal{D}_{obs}$: $$\mathcal{D}_{obs} = \{(X^{t_i^{pre}}_i, X^{t_i^{post}}_i, A^{t_i}_i)\}_{i=1}^n.$$ 

% $P^{\mathcal{D}_{{train}}}_{X^t, A^t, Y^t}$, $P^{\mathcal{D}_{{obs}}}_{X^t, A^t, Y^t}$

We additionally have access to a treatment-free training population with measurements and outcomes recorded longitudinally. This is our dataset for training,~$\mathcal{D}_{train}$.~We can write this for some $N$ patients as $$\mathcal{D}_{train} = \{(X_i^{t_i^1},X_i^{t^2_i},\ldots, Y^{t^1_i}_i,Y^{t^2_i}_i, \ldots)\}_{i=1}^N.$$ The union of these two~is~a~combined dataset for the surrogate set up. It is,~in~fact, a rather regular occurrence to have a retrospective cohort that matches $\mathcal{D}_{{train}}$ with patient data and long-term outcomes and new short term treated data without long term outcomes.\footnote{In multiple sclerosis, as another example, long-term natural-history cohorts link baseline MRI to disability outcomes measured over 15–20 years \citep{fisniku2008disability}, but with the widespread adoption of disease-modifying therapies, comparable long-term outcome data under treatment are not yet available, even as pre- and post-treatment imaging is routinely collected.}  Our set up also enables the use of an externally pre-trained AI model applied to new treated observations. For notational convenience, we let the binary random variable $D = 1$ denote a sample from $\mathcal{D}_{{obs}}$ and $D = 0$ a sample from $\mathcal{D}_{{train}}$. 

Given $\mathcal{D}_{train}$ and $\mathcal{D}_{obs}$, our goal is to estimate a causal effect on $Y^{t^{end}}$ for some time $t^{end}$ in the future. We allow $t^{end}$ to vary by individuals as long as the intervals $t^{end}-t^{post}$, $t^{end}-t^{pre}$ are fixed. Following the potential outcomes framework \citep{rubin1974estimating, imbens2015causal}, we seek an average treatment effect (ATT) in the~treated~dataset~$\mathcal{D}_{obs}$:
\begin{align*}
\mbox{ATT} \equiv \mathbb{E}\left[ {Y}^{t^{end}}(1) -
    {Y}^{t^{end}}(0)\mid A^{t^{post}}=1, D=1 \right]
\end{align*} for potential outcome ${Y}^{t^{end}}(0)$ under no treatment by $t^{post}$, i.e. $A^{t^{post}}=0$, and not yet observed outcome ${Y}^{t^{end}}(1)$ under treatment given by $t^{post}$, i.e. $A^{t^{post}}=1$. The potential outcome ${Y}^{t^{end}}(0)$ is never observed in $\mathcal{D}_{obs}$.~Because~outcomes ${Y}^{t^{end}}(1)$ are also not observed in $\mathcal{D}_{\text{obs}}$, we cannot estimate treatment effects via covariate adjustment~with~a matched control group either. Thus, we use an~AI~surrogate. 

We remark that we can also refer to potential outcomes and to $t^{post}$ in the untreated dataset $\mathcal{D}_{train}$. Since $A_i^t=0$ for all $t$ and each $i$ in $\mathcal{D}_{train}$, fixing some time $t^{post}_i$, any observation can be written ${Y}^{t^k}_i(0)$ with ${Y}_i^{t^k}(1)$ not observed for any individual or any $t^k$. 

\subsection{Causal Effect Estimation with Before-and-after AI Surrogates}

Using an AI surrogate for our set up means that we study when a model trained in $\mathcal{D}_{train}$ to predict $Y^t$ from $X^t$ can be used as replacement for knowing $Y^{t^{end}}$ in the treated population ($\mathcal{D}_{obs}$) using measurements $X^{t^{pre}}, X^{t^{post}}$. The AI model replaces $Y^{t^{end}}(0), Y^{t^{end}}(1)$ with predictions. Extending our notation, our AI model is trained to be a time-aware predictor approximating \begin{align}\label{model}
    \mu(X, t^1 - t^0) = \mathbb{E}\left[Y^{t^1} \mid X^{t^0},  D=0\right].
\end{align} The model $\widehat{\mu}$ estimating ${\mu}$ is trained from measurements $X^{t^0}$ and $Y^{t^1}$ to predict the outcome at multiple intervals $(t^1-t^0)$ into the future. Statistically, this model approximates the expected value for the longitudinal outcome $Y^{t^1}$ given the inputs $X^{t^0}$ in $\mathcal{D}_{\text{train}}$. In our ECG-based heart failure risk example, this means training a model that takes as input an ECG $X^{t^0}$ at any time $t^0$ and predicts $(t^1-t^0)$ time steps into the future the risk of heart failure, $Y^{t^1}$, at $t^1$ across many $t^1, t^0$ pairs. If we seek to understand five year heart failure risk, the interval $(t^1-t^0)$ would be five years and $t^0, t^1$ would correspond to calendar measurement times.

We apply this model as follows. For each measurement pair $X^{t^{pre}}, X^{t^{post}}$ in the treated dataset $\mathcal{D}_{obs}$, we use the trained model to calculate AI surrogate endpoints as
\begin{align*}
    \widehat{\mu}(X^{t^{pre}}, t^{end}-t^{pre}) \\ \intertext{estimating the untreated outcome at time $t^{end}$, and} \\
    \widehat{\mu}(X^{t^{post}}, t^{end}-t^{post})
\end{align*} estimating the treated outcome at time $t^{end}$. For ECGs and heart failure, this means taking the pre-treatment measurement and predicting with the model enough time steps into the future to approximate risk at $t^{end}$ and for the post treatment measurement, taking the smaller time window from this measurement to the same calendar time. For before-and-after measurements taken two years apart, predicting five year heart failure risk predicts five years out from $X^{t^{post}}$ and seven years out from $X^{t^{pre}}$. In our framework, because we model outcomes over time, patients serve as their own controls.  The former prediction approximates the counterfactual trend. For a perfect model $\widehat{\mu} = \mu$, this approach identifies the treatment effect.

\section{Causal Identification with Before-and-after AI Surrogates}

\textbf{Proposition 1}. Under the assumptions delineated below (Assumptions 1-6), we can write ATT as 
\begin{align*}
    \mbox{ATT} \,=&\,\, \mathbb{E}[\mu(X^{t^{post}}, t^{end}- t^{post}) \nonumber \\&-
   \mu(X^{t^{pre}}, t^{end} - t^{pre}) \mid A^{t^{post}}=1, D=1]. 
\end{align*}

\subsubsection*{Standard Assumptions}
To identify the ATT from $\mathcal{D}_{\text{train}}$ and $\mathcal{D}_{\text{obs}}$ using the before-and-after surrogates, i.e. to prove Proposition 1, we make a series of~assumptions.~We begin with standard conditions required for~causal~inference \citep{imbens2015causal, rosenbaum1983central}. First, we need \textit{consistency}:\\

\vspace{-.1in}
\noindent
\textbf{Assumption 1} (Consistency). The observed outcome and measurement at any time point under the treatment received match the outcome should that treatment have~been~randomly assigned. Mathematically, for any $t$, 
\begin{center}
\begin{minipage}{0.8\columnwidth}
\centering
$Y^t=Y^t(a)$ and $X^t=X^t(a)$ for $A^{t^{post}}=a$.
\end{minipage}
\end{center}

\noindent
We also require the standard notion of \textit{positivity}.\\

\vspace{-.1in}
\noindent
\textbf{Assumption 2} (Positivity). There is a non-zero probability for every measurement in the treated dataset to have appeared in the training dataset: 

\begin{center}
\begin{minipage}{0.8\columnwidth}
\centering
For $t \in \{t^{pre}, t^{post}\}$ and $x \in \mathcal{D}_{obs}$, $P(D=0 \mid X^t =x) > 0.$ 
\end{minipage}
\end{center}

These are analogues of standard assumptions for identifying causal effects in traditional settings. Consistency is an untestable assumption. In our setting, consistency includes the relationship between observed and potential measurements $X^{t^{pre}}=X^{t^{pre}}(1)$,\, $X^{t^{post}}=X^{t^{post}}(1)$ and measurements under no treatment, i.e. $X^{t^{pre}}(0) ,\, X^{t^{post}}(0)$,~in~addition to potential outcomes. For either variable, consistency requires that treatments are well-defined, and that multiple treatments with differing effects or administration are not considered as one treatment. Positivity requires that every covariate value in $\mathcal{D}_{\text{obs}}$ has positive probability of appearing in $\mathcal{D}_{\text{train}}$. The positivity condition is testable by comparing the observed covariate distributions, and assessing the overlap of their supports. 

\subsubsection*{Assumptions on Training Data}
 
Beyond the~standard~conditions, we require new testable assumptions about the relationship between $X^t$ and $Y^t$ over time.~First,~we~desire that the endpoint is predictable from a single covariate measurement, i.e. having earlier covariate information must not be necessary to improve the prediction. More specifically, we write this as a Markovian assumption.\\

\vspace{-.1in}
\noindent
\textbf{Assumption 3} (Markov Covariate Sufficiency). Later covariate measurements subsume the information present about the outcome $Y^t$ present in previous measurements:

\begin{center}
\begin{minipage}{0.8\columnwidth}
\centering
For $d \in \{0, 1\}$ and $t'' > t' > t$, $Y^{t''} \perp X^{t} \mid X^{t'}, D=d$.

\end{minipage}
\end{center}

%\textcolor{red}{Geoff: this isn't correct for things that show up sometimes like afib. Is there anything I should rephrase cuz of this or make a note that there are cases that this doesn't hold, ect.}
{{Assumption 3 simply requires that the earlier measurement does not provide additional information about $Y^{t}$ over the later measurement.} Assumption 3 does not yet require that $X^t$ describes all changes in $Y^t$, but rather that later data subsumes earlier measurements in their predictive value for an even later $Y^t$. This assumption is most reasonable when $X^t$ is a high-dimensional diagnostic summary of the subject's health state, but not guaranteed; for instance, in cardiology, a current ECG~can~captures the electrical state of the heart sufficiently for predicting certain cardiovascular events, regardless of prior recordings, but not others (such as paroxysmal atrial fibrillation). {We can test this assumption by comparing whether a model built with two measurements $X^t, X^{t'}$ improves prediction of $Y^{t''}$ over using only $X^{t'}$.}} 

Our framework also uses modeled controls~rather~than matching similar individuals by estimating~their~counterfactual outcomes using the~model~$\mu$. This~has~benefit in that we need not assume unconfoundedness and can estimate individualized effects but relies on assuming we can model the outcome that would have occurred using a measurement prior to the treatment. We call this  condition \textit{counterfactual baseline sufficiency}.\\

\vspace{-.1in}
\noindent
\textbf{Assumption 4} (Counterfactual Baseline Sufficiency). We can predict what would have happened at $t^{end}$ without treatment from the pre-treatment measurement without knowing the counterfactual measurement at $t^{post}$:

\begin{center}
\begin{minipage}{0.8\columnwidth}
\centering
$Y^{t^{end}}(0) \perp X^{t^{post}}(0) \mid X^{t^{pre}}(0)$.
\end{minipage}
\end{center}

The Markovian assumption requires that~once~we~have later data, earlier data is either outdated or~redundant. The counterfactual baseline sufficiency assumption supposes that if nothing external happens between $t^{post}$ and $t^{pre}$, i.e. no treatments or affecting actions, then we can estimate the baseline from the earlier time point as well. Assumption 4 specifically supposes that earlier measurements predict the outcome just as well as the later data if the short~trajectory~between them is intervention free, and the assumption only considers this case of no intervention between the time points $t^{pre}$, $t^{post}$. For reasonably nearby $t^{pre}$ and $t^{post}$, this is simply assuming a time aware model using $X^t$ predicts $Y^{t^{end}}$ well at either time point. {We can test Assumption 4 as well by comparing, for $t''> t' > t$, where $t'-t$ is small, whether a model built with two measurements $X^t, X^{t'}$ improves prediction of $Y^{t''}$ now over using only $X^{t}$.} \\

\vspace{-.1in}
\subsubsection*{Untestable Assumptions for AI Surrogacy} Additionally, for the AI model predictions to serve as valid surrogates for the unobserved outcome $Y^{t}$, we need the AI predictions to serve as a surrogate for the treatment effect on the endpoint. In our set up, this amounts to two assumptions. First, we need \textit{comparability}.\\

\vspace{-.1in}
\noindent \textbf{Assumption~5}~(Comparability).~The~relationship~between~the endpoint and the measurement in any intervention-free trajectory is independent of whether the data are from $\mathcal{D}_{obs}$ or $\mathcal{D}_{train}$. Let $t < t'$, then

\begin{center}
\begin{minipage}{0.8\columnwidth}
\centering
$Y^{t'}(0) \perp D \mid X^{t}(0)$.
\end{minipage}
\end{center} 

\noindent Comparability implies that for treatment free trajectories, once we know the value of the measurement $X^t$, there is no additional data about an outcome $Y^{t'}$ in knowing which dataset $D$ the measurement comes from. Assumption 5 allows a model trained to learn the relationship between $X^t$ and $Y^t$ in $\mathcal{D}_{train}$ to translate to $\mathcal{D}_{obs}$ in the absence of interventions. The assumption requires only that the intervention-free relationship between $X^t$ and $Y^{t}$ extends from $\mathcal{D}_{train}$ to $\mathcal{D}_{obs}$. 

To model treatment effects, we need that after treatment, i.e. at $t\geq t^{post}$ and $ A^{t^{post}}=1$, once there are no further interventions, for the relationship between $X^t$ and $Y^t$ to be the same relationship learned from the training data. This requires an additional assumption of \textit{surrogacy}. \\

\vspace{-.1in}
\noindent 
\textbf{Assumption 6} (Statistical Surrogacy). For any $t$ with $t'> t\geq t^{post}$ and for any  $A^{t^{post}} \in \{0,1\}$, whether the data are from $\mathcal{D}_{obs}$ or $\mathcal{D}_{train}$ (treated or untreated) does not provide more information on the endpoint after knowing $X^t$, i.e. 
\begin{center}
\begin{minipage}{0.8\columnwidth}
\centering
$Y^{t'} \perp D \mid X^{t}$.
\end{minipage}
\end{center}

Positivity guarantees any $X^{t^{post}}$ in $\mathcal{D}_{obs}$ could appear in $\mathcal{D}_{train}$, a prerequisite for this assumption. Comparability guarantees that $\mathcal{D}_{train}$ and $\mathcal{D}_{obs}$ do not at baseline have differing relationships between $X^t$ and $Y^{t}$. Surrogacy further guarantees that the relationship between $X^t$ and $Y^{t}$ is unchanged between the datasets after the treatment occurs, i.e. by the treatment. 

We note that surrogacy is much stronger than $X^t$ being highly associated with $Y^{t}$ in general. We need $X^{t^{post}}$ to capture the changes that $A^{t^{post}}$ inflicts on $Y^{t^{end}}$, so that observing $X^{t^{post}}$ is sufficient to observe the effect of $A^{t^{post}}$. The requires that the mechanism of action of $A^{t^{post}}$ on $Y^{t}$ shows up as a feature in $X^{t^{post}}$.\footnote{Note that we also implicitly require in selecting $t^{post}$ that the measurement is taken at a late enough time to satisfy Assumption 6 since the effect of $A^{t}$ on $X^t$ is likely not be immediately observable even if much faster than on $Y^{t^{end}}$.} If AI simply automates the collection of an endpoint, then $X^t$ would contain such features, and surrogacy will be satisfied, i.e. accurate pathology based models can be surrogates for pathological endpoints. AI predictions may also be a good endpoint when the mechanism of treatment effect appears in the measurement even beyond automation. {For ECGs, AI~would~be~a~good~surrogate for an arrhythmogenic endpoint or disease with mechanisms appearing on ECG, including heart failure if there are no treatment mechanisms missed completely by some signal in the ECG reading.}~Surrogacy~is~an~untestable assumption as it is a property on the treatment mechanism and must be a priori known not computed from $\mathcal{D}_{obs}$, which lacks endpoints $Y^{t^{end}}$. 

Finally, note that, since the AI surrogate model is trained on $\mathcal{D}_{\text{train}}$ but used for inference in $\mathcal{D}_{\text{obs}}$, we additionally require in comparability and surrogacy that the two datasets have the same relationship between measurements and endpoints in another sense. We cannot have that $X^t$ is a strong surrogate but that the AI model we use never sees examples of the specific mechanistic pathway of that surrogacy. The relationship learned by the AI model trained in $\mathcal{D}_{train}$ must be that which captures treatment effects on $X^t$. The assumptions capture that the relationship between the random variables $X^t, Y^t$ must satisfy surrogacy and be present in $\mathcal{D}_{train}$ and $\mathcal{D}_{obs}$.

Assumptions 1-6 give that our framework identifies true ATT with a perfect model for $\mu$ (Appendix \ref{app:proofs}). When Assumptions 1-6 are satisfied, AI surrogates give us treatment effects from high-dimensional data. When any are absent, this effect identification does not hold. 

\section{Estimation and Inference of Treatment Effects}

We perform inference using the AI surrogate by computing predictions on each individual from the observed data $\mathcal{D}_{obs}$, i.e. we compute AI predicted endpoints for each patient $\widehat{Y}^{t^{end}_i}_i(1), \widehat{Y}^{t^{end}_i}_i(0)$, which we define as
\begin{align}
     \widehat{Y}_i^{t^{end}_i}(0) =  \widehat{\mu}(X_i^{t_i^{pre}}, t_i^{end}-t_i^{pre}) \\ \intertext{and}\nonumber \\
   \widehat{Y}_i^{t^{end}_i}(1) =\widehat{\mu}(X_i^{t_i^{post}}, t_i^{end}-t_i^{post}). 
\end{align}

Under our assumptions and a perfect model,~our~estimator for ATT would be the average of the difference of the two model predictions for each data point in $\mathcal{D}_{obs}$: \begin{align}\label{ATT_hat}
   \widehat{\text{ATT}} :=\frac{1}{n}\sum_{i=1}^n\left( \widehat{Y}^{t_i^{end}}_i(1) -
    \widehat{Y}^{t_i^{end}}_i(0) \right).
\end{align} The AI model predictions serve as the surrogate outcomes. Estimation with any model is, however, in general biased by model errors. In particular, using the linearity~of~expectation, we can write $\mbox{ATT} - \mathbb{E}[\widehat{\mbox{ATT}}]$ as 
\begin{align}
&=
\underbrace{\mathbb{E}\left[
Y_i^{t_i^{end}}(1)
-
\widehat{Y}_i^{t^{end}}(1)
\mid A^{t^{post}}=1, D=1
\right]}_{\text{How wrong we are on }Y^{t^{end}}(1)} 
\\
&\quad -
\underbrace{\mathbb{E}\left[
Y_i^{t_i^{end}}(0)
-
\widehat{Y}_i^{t^{end}}(0)
\mid A^{t^{post}}=1, D=1
\right].}_{\text{How wrong we are on }Y^{t^{end}}(0)}
\end{align}
If we are equally wrong at predicting the outcome with the AI surrogate model on average on both before and after treatment distributions, these biases will cancel. Thus, with additional assumptions on the distribution of data in $\mathcal{D}_{obs}, \mathcal{D}_{train}$ and on model bias, and by Assumptions 1-6, we can selectively use this estimator for valid inference.\\

\vspace{-.1in}
\noindent
\textbf{Proposition 2.} \label{base_estimator} Define the estimator $\widehat{ATT}$ as in Equation \ref{ATT_hat}. Assume Assumptions 1-6 and that the model bias $\mathbb{E}[Y^{t+t'} - \hat{\mu}(X^t, t')]$ is constant in $t'$ and between samples with \textit{unobserved} labels $(X^{t^{pre}}, Y^{t^{pre}+t'}(0))$ and $(X^{t^{post}}, Y^{t^{post}+t'}(1))$. Define sample variance $\displaystyle \widehat{\sigma}_{ATT}^2 = \frac{1}{n-1}\sum_{i=1}^n(\Delta_i - \bar{\Delta})^2$ for $\Delta_i =\widehat{Y}_i^{t_i^{end}}(1) - \widehat{Y}_i^{t_i^{end}}(0)$ with mean $\bar{\Delta}$. Then with $z_{1-\alpha/2}$ denoting the $1-\alpha/2$ quantile of the standard normal distribution, the interval 
\begin{align}\label{estimator1}
        \widehat{ATT} \pm z_{1-\alpha/2} \sqrt{\frac{\widehat{\sigma}_{ATT}^2}{n}}
    \end{align} has asymptotic coverage of ATT at level $1-\alpha$ assuming finite true variance and consistent variance estimators.

The proof is provided in Appendix \ref{app:estimators_proofs}.~In~general,~unfortunately, how wrong we are at predicting the outcome after treatment will differ from how wrong we are at predicting it prior as the measurements may differ in distribution (shifting model errors) and, moreover, surrogacy may not hold. We also predict~farther~into the future from $X^{t^{pre}}$ than from $X^{t^{post}}$, which may change model bias. Without perfect satisfaction of the above assumptions, inference with the above estimator is invalid. 

Suppose now that we do not have perfect surrogacy, i.e. no Assumption 6, but we do have some data on the true observed outcomes in the treated population $\mathcal{D}_{obs}$. With a bias corrected estimator reminiscent of prediction powered inference and its predecessors \citep{ppi, neyman1938contribution}, we can correct for differences in model errors and failures in surrogacy at once. We use access to~a~limited~sample~of~data on the true outcomes $Y^{t^{end}}(1)$ in the treated population $\mathcal{D}_{obs}$ and a sample of untreated individuals with endpoints, i.e. from $\mathcal{D}_{train}$. With both imperfect models and imperfect surrogacy, this second estimator and access to a small labeled treated sample allows us to still estimate an unbiased average treatment effect. In practice, since surrogacy is untestable, this estimator is preferable.\\

\vspace{-.05in}
\noindent 
\textbf{Proposition 3.} \label{ppi2} Assume Assumptions 1-5. Assume additionally that $(X^{t^{pre}}, Y^{t^{end}}(0))$ can be viewed as a random sample from the same underlying distribution as equally spaced observations $(X^{t^{0}}, Y^{t^{k}})$ in $\mathcal{D}_{train}$. Imagine we have labels $Y^{t^{end}}(1)$ for a size $n_L$ subset of $\mathcal{D}_{obs}$, which has $n$ total observations. Calculate for each individual in $\mathcal{D}_{obs}$:
\begin{align*}
    \widehat{Y}_i^{t_i^{end}}(0)  = \widehat{\mu}(X_i^{t_i^{pre}}, t_i^{end}-t_i^{pre}) \\
    \widehat{Y}_i^{t_i^{end}}(1)  = \widehat{\mu}(X_i^{t_i^{post}}, t_i^{end}-t_i^{post}).
\end{align*} Take a new independent sample from the same distribution as $\mathcal{D}_{train}$ not used in training of size $N$. Calculate for these individuals with the same time gap $k$ between $X^{t^{pre}}$ and $Y^{t^{end}}$ as between $t^{pre}$ and $t^{end}$:
\begin{align*}
    \widehat{Y}_j^{t_i^k}  = \widehat{\mu}(X_j^{t_i^0}, t_j^k-t_j^0).
\end{align*}
Define the estimator $\widehat{ATT}^{PPI}$ to be
    \begin{align}\label{ppi_estimator_eqn}
        &\frac{1}{n-n_L}\sum_{i=n_L+1}^{n} (\widehat{Y}_i^{t_i^{end}}(1) - \widehat{Y}_i^{t_i^{end}}(0)) \nonumber \\&+  \frac{1}{n_L}\sum_{i=1}^{n_L} (Y^{t_i^{end}}_i(1) - \widehat{Y}_i^{t_i^{end}}(1)) \nonumber -  \frac{1}{N}\sum_{j=1}^N (Y_j^{t_j^k} - \widehat{Y}_j^{t_j^k}) 
    \end{align} 
    with sample variances \begin{align*}
        \widehat{\sigma}_{ATT}^2 &=\frac{1}{n-n_L-1}\sum_{i=n_L+1}^{n}({\Delta_{ATT}}_i - {\bar{\Delta}_{ATT}})^2 \\
        \widehat{\sigma}_{post}^2 &=\frac{1}{n_L-1}\sum_{i=1}^{n_L}({\Delta_{post}}_i - {\bar{\Delta}_{post}})^2 \\
        \widehat{\sigma}_{control}^2 &=\frac{1}{N-1}\sum_{j=1}^N({\Delta_{control}}_j - {\bar{\Delta}_{control}})^2
    \end{align*} 
    for ${\Delta_{ATT}}_i =\widehat{Y}_i^{t_i^{end}}(1) - \widehat{Y}_i^{t_i^{end}}(0)$ with mean $\bar{\Delta}_{ATT}$, ${\Delta_{post}}_i ={Y}_i^{t_i^{end}}(1) - \widehat{Y}_i^{t_i^{end}}(1)$ with mean $\bar{\Delta}_{post}$, and ${\Delta_{control}}_j ={Y}_j^{t_j^{k}} - \widehat{Y}_j^{t_j^{k}}$ with mean $\bar{\Delta}_{control}$, respectively.  Then, for $z_{1-\alpha/2}$, the interval created \begin{align}
        \widehat{ATT}^{PPI} \pm z_{1-\alpha/2} \sqrt{\frac{\widehat{\sigma}_{ATT}^2}{n-n_L} + \frac{\widehat{\sigma}_{post}^2}{n_L} + \frac{\widehat{\sigma}_{control}^2}{N}}
    \end{align} has asymptotic coverage of ATT at level $1-\alpha$ assuming finite true variance and consistent variance estimators.
    
We prove this result in Appendix \ref{app:estimators_proofs}. We further relax distributional assumptions and suggest even more general estimators with a covariance shift in Appendix \ref{app:estimators}. The debias corrections we propose use a small set of labels on the treated group and a set of untreated population labels. If the model was trained by the same researchers, a holdout from $\mathcal{D}_{train}$ can be used for the untreated group. If not, any untreated patients may be used without needed to match any characteristics of these patients using the covariate shift results. The patients must be an independent sample and not used in training the AI model.

\section{Heterogeneous Effect Estimation}

Another advantage of the before-and-after AI surrogate approach is that using individuals as their own controls suggests an identification result for heterogeneous effects as well. We define $CATT(x)$ as
\begin{align*}
\mathbb{E}\left[Y^{t^{end}}(1) - Y^{t^{end}}(0) \mid X^{t^{pre}}=x, A^{t^{post}}=1, D=1 \right]
\end{align*} the average treatment effect for a treated individual conditional on the pre-treatment measurement. We can also identify $CATT(x)$ with our approach.\\

\vspace{-.05in}
\noindent 
\textbf{Proposition 4}. Under the same assumptions as above (Assumptions 1-6), we can write CATT as 
\begin{align}
&\mathbb{E}[
\mu(X^{t^{post}}, t^{end}-t^{post})
- \nonumber \\
& 
\mu(x, t^{end}-t^{pre})
\mid
X^{t^{pre}}=x,\,
A^{t^{post}}=1,\,
D=1
].
\end{align}
If we can average over individuals with the same measurement $X^{t^{pre}}$ alongside labels, we would have an analogous estimator for CATT as above. However, in the absence of this, heterogeneous effect estimation will be biased by model and surrogacy biases. We can group individuals by heterogeneity in low-dimensional feature sets $\widetilde{X}$ (i.e. all older Hispanic females or all men under 40 with diabetes) as long as there is sufficient count. This produces average heterogeneous treatment effects with analogous estimators to the above conditioned on $\widetilde{X}$, but we cannot obtain unbiased individual effects. 

Despite not being able to correct for individual bias, in clinical and pharmacological practice, AI based surrogates can still serve as useful tools for monitoring individual response in conjunction with additional confirmatory tests. Even in the absence of true surrogacy, we estimate a treatment effect on the AI surrogate model $\widehat{\mu}$. If we know an AI ECG model predicts heart failure through specific mechanisms based on the training data, we can estimate the effect of a treatment on those mechanisms for patients. In settings of pharmacovigilance with known negative effects to avoid, we might use a trained AI ECG aware of the mechanisms of these effects to detect them early.\footnote{Pharmacovigilance models are often built to detect specific mechanisms i.e. \cite{prifti2021deep, schubert2026deep}.} Critically, however, we lack the ability to predict risk with this model of unknown mechanisms or for outcomes that may have mechanisms missed by the AI surrogate. This requires an average over labeled data with the true outcomes $Y^t$ and cannot be computed for individuals without unknown bias.

\section{Experiments}

\subsection{Synthetic Experiments}

We demonstrate our method with a simple synthetic simulation study comparing the strength of a surrogate and the quantity of labeled data. In these experiments, we model a longitudinal outcome $Y^t$ as a non-linear function of two latent variables $Z_1^t, Z_2^t$ that progresses auto-regressively starting from random noise, i.e. $Z_1^0, Z_2^0\sim \mathcal{N}(0,1)$. We consider both a linear and non-linear process for $Z_1^t, Z_2^t$:
\begin{align*}
    &\text{\textit{Linear: }} Z_j^t = \rho_Z * Z_j^{t-1} \\
    &\text{\textit{Non-linear: }} Z_j^t = \rho_Z * Z_j^{t-1} + \beta * \tanh(Z_j^{t-1}). 
\end{align*} For the linear process, we set $\rho_z=1.01$ and for the non-linear process, $\rho_z=0.85, \beta=0.25$. A treatment bump to $Z^t_1$ is added once during process construction as
\begin{align*}
    Z^{t^{post}}_1(1) = Z^{t^{post}}_1(0) + 0.5.
\end{align*} Treatment leaves $Z^t_2$ unchanged, i.e. $Z^t_2(1) = Z^t_2(0)$ for all $t$. Each potential outcome $Y^t(a)$ is written in terms of $Z_1^t(a), Z_2^t(a)$, explicitly we choose
\begin{align*}
    Y^t(a) = 1 + Z_1^t(a) + Z_2^t(a) + 0.5 * (Z_1^t(a) + Z_2^t(a))^2 + \varepsilon^t
\end{align*} for $\varepsilon^t \sim \mathcal{N}(0, 0.05)$. The data $X^t$ is then a two dimensional variable created as a function of $Z_1^t, Z_2^t$ depending on a parameters $\rho_{X_1}, \rho_{X_2}$ governing the degree of surrogacy:
\begin{align*}
    X_j^t = \rho_{X_j} Z_j^t +  \sqrt(1-\rho_{X_j}^2) \varepsilon^t
\end{align*} for noise $\varepsilon \sim \mathcal{N}(0,1)$.

We consider surrogates with permutations of $\rho_{X_1}=1, 0.85, 0.5$ and $\rho_{X_2}=1, 0.85, 0.5$. This allows us to evaluate predictions that are highly correlated with the outcome but poor surrogates, i.e. with $\rho_{X_1}<1$ but $\rho_{X_2}=1$ in contrast to the effect of a poor surrogate i.e. $\rho_{X_1}=\rho_{X_2}=0.5$. For each data generation process, we sample 20,000 points to build a training dataset of entirely untreated trajectories. A 20\% validation hold out is used to determine model performance. We build a sample of 5,000 treated observational points ($\mathcal{D}_{obs}$) and 5,000 held out untreated points (from $\mathcal{D}_{train}$) for inference. We consider $t^{pre},t^{post}$ as two steps apart, and the endpoint $t^{end}$ five time steps from $t^{post}$. This simulation is repeated 50 times for each $\rho_{X_1}, \rho_{X_2}$ permutation for each of the linear and non-linear processes. First, we take 5\% of this data as \textit{labeled} and report the bias, interval size, and the coverage of the 95\% confidence interval created using Equation \ref{ppi_estimator_eqn} (PPI) as compared to the naive AI surrogate estimator with all data (Naive, Equation \ref{estimator1}) and difference in means (DIM) estimators with the same 5\% of data (Table \ref{tab:simulations_surrogacy}). We see that the naive AI surrogate gives a minimal bias in estimate of ATT when $\rho_{X_1}=1$ as expected. As the value of $\rho_{X_1}$ degrades, the bias grows but the PPI correction debiases the effect. The PPI interval size decreases with both improved surrogates and improved model performance on the overall outcome $Y^{t^{end}}$, i.e. as $\rho_{X_1}, \rho_{X_2}$ grow toward 1.

\begin{table*}[h!]
\centering
\tiny
\setlength{\tabcolsep}{5pt}
\begin{tabular}{cc|cc|ccc|c|cc}
\toprule
& & \multicolumn{2}{c|}{Model Prediction} & \multicolumn{3}{c|}{PPI} & Naive & \multicolumn{2}{c}{DIM} \\
$\rho_{X1}$ & $\rho_{X2}$ & MAE & $R^2$ & Coverage & Bias & Interval & Bias & Bias & Interval \\
\midrule
\multicolumn{10}{l}{\textbf{Linear DGP}} \\
\addlinespace[2pt]
0.5 & 0.5 & 1.380 (0.003) & 0.169 (0.002) & 90.0\% & 0.015 (0.035) & 0.811 (0.007) & -0.533 (0.003) & -0.038 (0.033) & 0.891 (0.009) \\
0.5 & 1 & 1.028 (0.002) & 0.514 (0.003) & 92.0\% & 0.027 (0.024) & 0.631 (0.005) & -0.529 (0.003) & 0.011 (0.032) & 0.886 (0.009) \\
0.85 & 0.85 & 0.882 (0.002) & 0.637 (0.002) & 98.0\% & -0.031 (0.017) & 0.549 (0.005) & -0.213 (0.004) & 0.015 (0.032) & 0.891 (0.009) \\
0.85 & 1 & 0.634 (0.002) & 0.807 (0.001) & 90.0\% & -0.009 (0.019) & 0.401 (0.003) & -0.217 (0.003) & -0.036 (0.036) & 0.902 (0.009) \\
1 & 0.5 & 1.030 (0.003) & 0.515 (0.003) & 94.0\% & 0.035 (0.021) & 0.625 (0.006) & -0.010 (0.003) & 0.024 (0.032) & 0.904 (0.009) \\
1 & 0.85 & 0.633 (0.002) & 0.807 (0.001) & 96.0\% & -0.015 (0.015) & 0.404 (0.005) & -0.010 (0.003) & -0.003 (0.041) & 0.896 (0.008) \\
1 & 1 & 0.068 (0.000) & 0.996 (0.000) & 98.0\% & -0.000 (0.003) & 0.095 (0.004) & -0.005 (0.001) & 0.031 (0.029) & 0.890 (0.009) \\
\addlinespace[3pt]
\multicolumn{10}{l}{\textbf{Non Linear DGP}} \\
\addlinespace[2pt]
0.5 & 0.5 & 1.332 (0.003) & 0.178 (0.002) & 90.0\% & 0.016 (0.029) & 0.712 (0.004) & -0.449 (0.003) & -0.028 (0.028) & 0.780 (0.005) \\
0.5 & 1 & 1.010 (0.002) & 0.504 (0.002) & 92.0\% & 0.020 (0.018) & 0.543 (0.003) & -0.437 (0.002) & -0.001 (0.028) & 0.779 (0.005) \\
0.85 & 0.85 & 0.833 (0.002) & 0.641 (0.002) & 90.0\% & -0.018 (0.017) & 0.476 (0.003) & -0.174 (0.003) & 0.012 (0.028) & 0.778 (0.004) \\
0.85 & 1 & 0.589 (0.002) & 0.808 (0.001) & 92.0\% & 0.001 (0.015) & 0.339 (0.002) & -0.166 (0.003) & -0.033 (0.029) & 0.789 (0.005) \\
1 & 0.5 & 1.011 (0.002) & 0.506 (0.002) & 100.0\% & 0.027 (0.018) & 0.572 (0.003) & -0.013 (0.003) & 0.019 (0.028) & 0.788 (0.005) \\
1 & 0.85 & 0.589 (0.001) & 0.808 (0.001) & 96.0\% & -0.014 (0.013) & 0.362 (0.002) & -0.013 (0.003) & -0.023 (0.037) & 0.781 (0.004) \\
1 & 1 & 0.057 (0.000) & 0.998 (0.000) & 100.0\% & -0.001 (0.001) & 0.053 (0.001) & 0.001 (0.000) & 0.035 (0.026) & 0.784 (0.004) \\
\addlinespace[3pt]
\bottomrule
\end{tabular}
\caption{Coverage, bias, and confidence interval width by data generation process across linear, non-linear and $\rho_{X_1}, \rho_{X_2}$ to vary surrogacy. We use 5\% of the data as labeled. Bias and interval width are reported as mean with standard error across simulation runs.}
\label{tab:simulations_surrogacy}
\end{table*}

We next consider when in the case of having a poor surrogate we can still make conclusions about ATT. We fix a poor surrogate in both the linear and non-linear processes with $\rho_{X1}=0.5, \rho_{X1}=1.0$ and vary proportions of the labeled data (Table \ref{tab:simulations_vary_data}). We do the same for a great surrogate with $\rho_{X1}=1.0, \rho_{X1}=1.0$ (Table \ref{tab:simulations_vary_data_good}). Results for each labeled fraction are averaged over 50 trials. At less than 100\% of data being labeled, even with a poor surrogate, using the PPI debiased AI surrogate has a narrower confidence interval than using difference in means. With a great surrogate, inference has much more precision than naively using only the labeled data. With 100\% of data labeled, we do not need to use the AI
surrogate

In the appendix, we see consistent results in a set of experiments where the treatment assignment and outcome are confounded (Appendix \ref{app:exps}). There are extensions of prediction powered inference (PPI++, RePPI) that scale the estimator terms toward or away from predictions by minimizing the variance \citep{angelopoulos2023ppi++, miao2025assumption, van2026calibeating, ji2025predictionssurrogatesrevisitingsurrogate}. These methods may further improve prediction intervals and the robustness to poor surrogates for before-and-after AI surrogates as well.

\begin{table*}[h!]
\centering
\tiny
\setlength{\tabcolsep}{4pt}
\begin{tabular}{lc|ccc|ccc|c}
\toprule
& & \multicolumn{3}{c|}{PPI} & \multicolumn{3}{c|}{DIM} & Naive \\
DGP & Labeled data & Coverage & Bias & Interval & Coverage & Bias & Interval & Bias \\
\midrule
Linear & 1\% & 98.0\% & 0.012 (0.046) & 1.407 (0.031) & 92.0\% & 0.048 (0.070) & 1.987 (0.035) & -0.526 (0.002) \\
 & 10\% & 94.0\% & -0.026 (0.015) & 0.446 (0.003) & 94.0\% & -0.017 (0.027) & 0.629 (0.004) & -0.527 (0.002) \\
 & 50\% & 98.0\% & -0.004 (0.006) & 0.214 (0.001) & 100.0\% & -0.008 (0.009) & 0.281 (0.001) & -0.527 (0.002) \\
 & 100\% &  & &  & 94.0\% & -0.006 (0.007) & 0.199 (0.001) & -0.525 (0.003) \\
\addlinespace[3pt]
\midrule
\addlinespace[3pt]
Non-linear & 1\% & 98.0\% & 0.033 (0.042) & 1.195 (0.016) & 96.0\% & 0.036 (0.060) & 1.766 (0.018) & -0.433 (0.002) \\
 & 10\% & 96.0\% & -0.026 (0.013) & 0.383 (0.002) & 92.0\% & -0.023 (0.023) & 0.550 (0.002) & -0.435 (0.002) \\
 & 50\% & 100.0\% & -0.005 (0.006) & 0.186 (0.000) & 96.0\% & -0.008 (0.008) & 0.246 (0.001) & -0.437 (0.002) \\
 & 100\% & &  & & 90.0\% & -0.005 (0.007) & 0.175 (0.000) & -0.433 (0.003) \\
\bottomrule
\end{tabular}
\caption{Coverage, bias, and confidence interval width by labeled data fraction for a fixed poor surrogate $\rho_{X_1}=0.5, \rho_{X_2}=1.0$. Bias and interval width are reported as mean with standard error across simulation runs. Naive bias uses 100\% of the data as AI surrogates.}
\label{tab:simulations_vary_data}
\end{table*}

\begin{table*}[h!]
\centering
\tiny
\setlength{\tabcolsep}{4pt}
\begin{tabular}{lc|ccc|ccc|c}
\toprule
& & \multicolumn{3}{c|}{PPI} & \multicolumn{3}{c|}{DIM} & Naive \\
DGP & Labeled data & Coverage & Bias & Interval & Coverage & Bias & Interval & Bias \\
\midrule
Linear & 1\% & 98.0\% & -0.018 (0.006) & 0.146 (0.014) & 92.0\% & -0.035 (0.076) & 1.998 (0.040) & -0.004 (0.001) \\
 & 10\% & 100.0\% & 0.002 (0.002) & 0.074 (0.002) & 94.0\% & -0.034 (0.021) & 0.630 (0.005) & -0.005 (0.001) \\
 & 50\% & 98.0\% & 0.001 (0.002) & 0.069 (0.000) & 90.0\% & -0.008 (0.011) & 0.282 (0.001) & -0.005 (0.001) \\
 & 100\% & &  & & 96.0\% & -0.002 (0.007) & 0.199 (0.000) & -0.006 (0.001) \\
\addlinespace[3pt]
\midrule
\addlinespace[3pt]
Non-linear & 1\% & 96.0\% & -0.003 (0.002) & 0.083 (0.003) & 96.0\% & -0.005 (0.065) & 1.772 (0.021) & 0.001 (0.000) \\
 & 10\% & 100.0\% & 0.001 (0.001) & 0.047 (0.000) & 92.0\% & -0.032 (0.018) & 0.553 (0.003) & 0.002 (0.000) \\
 & 50\% & 100.0\% & -0.000 (0.001) & 0.054 (0.000) & 92.0\% & -0.002 (0.010) & 0.247 (0.001) & 0.002 (0.000) \\
 & 100\% & &  &  & 98.0\% & -0.000 (0.005) & 0.175 (0.000) & 0.002 (0.000) \\
\bottomrule
\end{tabular}
\caption{Coverage, bias, and confidence interval width by labeled data fraction for a fixed great surrogate $\rho_{X_1}=1.0, \rho_{X_2}=1.0$. Bias and interval width are reported as mean with standard error across simulation runs. Naive bias uses 100\% of the data as AI surrogates.}
\label{tab:simulations_vary_data_good}
\end{table*}

\subsection{Real World Treatment Effect Estimation in Cardio-Oncology}

We also consider a real data set up using the ECGs from the University of California, San Francisco to detect the cardiovascular effect of anthracycline and trastuzumab therapies in breast cancer patients. Both treatments are known to cause an increase in heart failure risk, for example recently in \cite{vo2024long}, but the mechanisms of heart failure risk are thought to differ between the two therapies \citep{zhang2022cardiac, camilli2024anthracycline}. Anthracyclines cause structural damage that increases with increasing dose and is largely irreversible. Trastuzumab causes a dose-independent risk increase due to blocking signaling pathways, and this risk increase is thought to be reversible. These two treatments have also both been used for decades, meaning we have long term outcomes to compare the use of an AI surrogate to. 

\begin{figure}[h!]
    \centering
    \includegraphics[width=\linewidth]{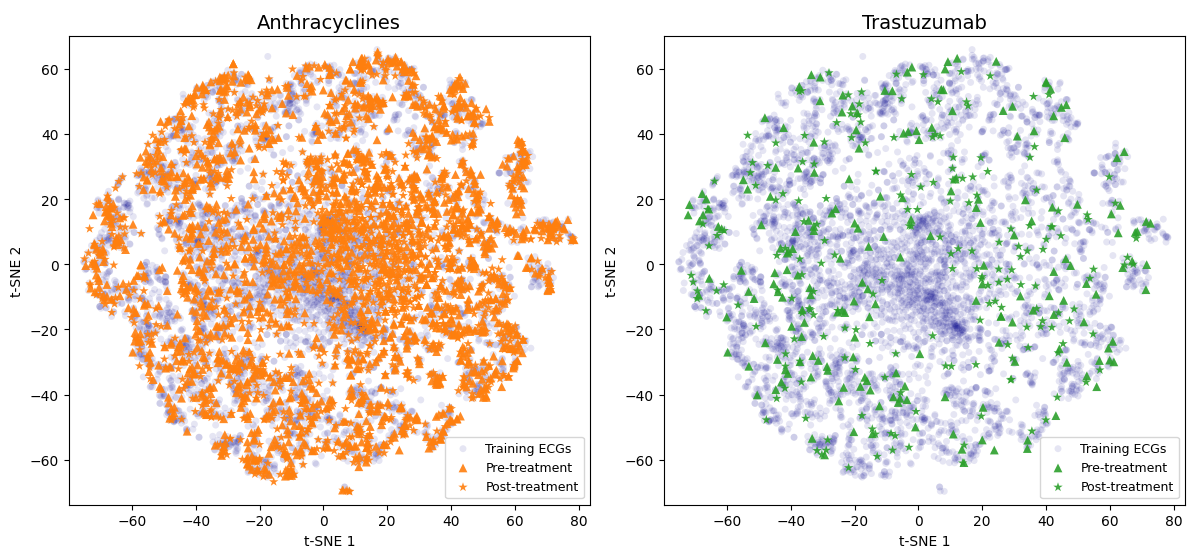}
    \caption{\textbf{Positivity check in cancer data}. {ECGs for cancer patients treated with anthracyclines and trastuzumab have large distributional overlap with the general model training population using dimensionality reduction visualization techniques.}}
    \label{fig:pos_anth_traz}
\end{figure}

For these analyses, an AI model is trained, without anthracycline and trastuzumab treated patients, to predict heart failure over time for 15 years. The AUROC\footnote{AUROC is the Area Under the Receiver Operating Characteristic which scores the model's precision and recall under all possible thresholds. A score of 0.5 indicates random performance and a score of 1.0 is perfect.} of this model averaged over 15 years is 0.847 (SE 0.00276). We see sufficient overlap in the distribution of ECGs using dimensionality reduced model embeddings suggesting that positivity is met (Figure \ref{fig:pos_anth_traz}). We also have that a model trained to predict heart failure with the addition of either an earlier or later ECG to assess for Markov covariate sufficiency and counterfactual baseline sufficiency, respectively, have minimal differences in predictive performance. The average AUROC difference across years is 0.016 (SE 0.005) compared to the model adding earlier data and 0.013 (SE 0.006) compared to the model adding later data. We defer further details on these assumption checks to Appendix \ref{app:exps}.	

\begin{figure}[h!]
    \centering
    \includegraphics[width=\linewidth]{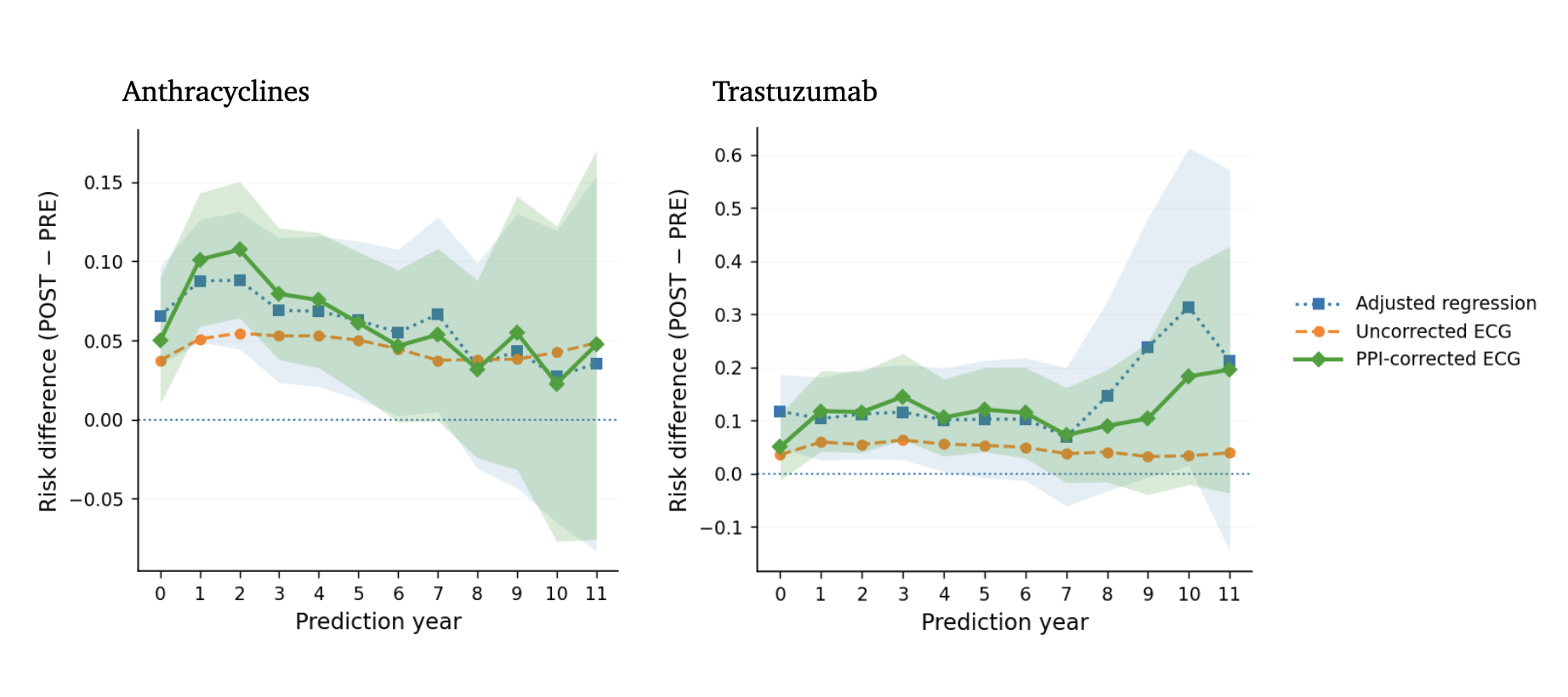}
    \caption{\textbf{Cardio-oncology experiments}. In breast cancer patients treated with anthracyclines (N=644) and trastuzumab (N=195), we achieve consistent results using available long-term real outcomes and covariate adjustment to using the model predictions as AI surrogates with a debias correction using 50\% of labels. We see that the AI surrogate prior to debias correction is a better surrogate in the case of anthracyclines than trastuzumab. Only years with at least 10 patients for label correction are plotted.}
    \label{fig:anth_traz}
\end{figure}

Taking the closest pre-treatment ECG then the latest ECG within three years of treatment for each treated patient, we calculate a before-and-after AI surrogate effect for anthracyclines and trastuzumab (Figure \ref{fig:anth_traz}). We consider only patients with breast cancer. We plot the naive surrogate (Equation \ref{estimator1}) and the debias correction (Equation \ref{ppi_estimator_eqn}) using labels on 50\% of the patients. In the appendix, we see similar results using fewer and more labels (Appendix \ref{app:exps}). We compare these AI predicted average treatment effect on heart failure risk to effects estimated using breast cancer patients treated with other therapies as controls and standard covariate adjustment. We regress controlling for other cancer treatments and standard cardiovascular risk factors (similar to \cite{vo2024long}). We see consistent results for heart failure risk in these cohorts over time for each treatment with the debiased AI surrogate. The naive surrogate better recovers anthracycline effects than trastuzumab effects. 

\section{Discussion}

Great AI surrogates would have clear advantages over traditional causal inference with real endpoints. The first is speed. If we can predict treatment effects using an AI model accurately, we reduce the need to wait for expensive or long-term outcomes. Second, as AI models are increasingly built as multi-modal and multi-outcome predictors, AI based surrogates, especially before-and-after AI surrogates, are particularly well suited for improving pharmacovigilance. Rather than waiting for unknown rare side effects to be observed across healthcare systems to generate testable hypotheses of unintended drug effects, an AI system could run on electronic medical records in the background. Such a collection of AI surrogates could detect changes in risk for a variety of clinical outcomes after new treatments automatically for treated patients. Our before-and-after AI surrogate framework does not require a control cohort, which means we do not need to understand confounders for each outcome before building such a system.  

Before-and-after AI surrogates compliment recent work where in unconfounded settings regression surrogates with cases and controls identify average treatment effects \citep{athey2025surrogate, ji2025predictionssurrogatesrevisitingsurrogate}.~The~framework~proposed by \cite{athey2025surrogate} can be viewed as proposing AI surrogates and requires the cross sectional versions of our untestable assumptions of surrogacy and comparability. We further detail their approach in Appendix \ref{related}. The before-and-after surrogate approach is broadly suitable when we cannot assume unconfoundedness and have access to longitudinal data, but both our and \cite{athey2025surrogate} frameworks will be necessary for the broad application and evaluation of AI based surrogates with \cite{athey2025surrogate} being particularly well suited to augmenting clinical trials and ours to understanding long term side effects without well-defined controls.~{The~estimator~proposed by~\cite{ji2025predictionssurrogatesrevisitingsurrogate}, also explored in Appendix \ref{related}, is a general set up for surrogate inference with predictions that includes treatment effect estimation while ours is a new treatment effect estimator. With \cite{ji2025predictionssurrogatesrevisitingsurrogate}, existing estimators are augmented using predictions and re-calibrated so that in the case that predictions are poorly calibrated PPI always improves over using only the labeled data (RePPI). Future work may be able to extend their calibration to our setting.}

Some AI based surrogate endpoints will automate extracting clinical data that is already known to be a surrogate. In this case, for features that AI can reliably measure well, AI could be expected to reduce bias in human measurement \citep{kadakia2026adoption}. Many pathology based-endpoints, like AIM-MASH, will likely have this nature. Known surrogates extracted with AI may reduce human error and facilitate trial speed. AI has additional advantages in being trained on high dimensional data and seeing what clinicians sometimes cannot. For instance, the features that AI uses to predict long term heart failure risk from ECGs are not entirely transparent. In our cardio-oncology case study, it is believed that anthracyclines and trastuzumab increase heart failure risk through different mechanisms \citep{zhang2022cardiac, camilli2024anthracycline}. Our AI ECG model appears better at capturing the irreversible oxidative stress mechanism of anthracyclines compared to the primarily asymptomatic reversible mechanism of trastuzumab. However, the debias correction gives valid inference in either case.

Still, the identification conditions required for the validity our approach reveal a need for great caution alongside excitement. Strong predictive performance of AI models does not mean a model is a good treatment effect estimator. The assumptions necessary for causal inference with AI models are strong. Surrogacy and comparability are untestable. Without a sample of labels on true outcomes, we cannot quantify the magnitude of bias caused by failure of these assumptions. Examples of where AI surrogates fail will abound just like with traditional biomarkers since the requirement of statistical surrogacy is strict and analogous. We remain excited that AI based surrogates can improve clinical monitoring and expedite research and pharmacovigilance, but stress that these assumptions require careful treatment before using AI surrogates in novel settings without labels.

\section{Additional Methods}

The formal proof of the methodology for our surrogate estimation framework appears in the appendix. We present details of our real world studies here with further details in Appendix \ref{app:exps}. All code for experiments was written in Python 3.11.13. The study protocol was approved by IRB 19-27430.

\subsection{Real ECG Data} 

We use a sample of 491,090 twelve lead ECG waveforms from UCSF obtained from 1980 up through 2021. ECGs were collected as a part of routine care using the MUSE software (GE Healthcare). ECGs were sampled at collection at either 500 or 250 Hz. All ECGs were down-sampled to 250 Hz and normalized using z-scores. We use accompanying electronic health record (EHR) data including any ICD diagnosis codes in conjunction with the UCSF cardiologist adjudicated MUSE diagnoses to create our disease labels. Labels are masked for training and inference following latest EHR entry. For model development, the data were split into 70\% training, 10\% validation for tuning and early stopping, and 20\% testing. We construct the semi-synthetic held out treated group from the testing cohort. All anthracycline and trastuzumab treated patients were separately held out prior to data splitting. 

\subsection{AI ECG Model}

Our AI ECG architecture has an initial convolution followed by a four stage shared residual neural network (ResNet) backbone with 3, 4, 6, and 3 residual blocks, respectively. Following the backbone, the model has a head for disease prediction. The network uses batch normalization, ReLU activation, and dropout throughout. Tabular covariates, including of age, race ethnicity, sex, and basic cardiovascular risk factors are concatenated to the ECG ResNet embedding and passed through a small multi-layer perceptron in the prediction head to produce an estimate for the hazard at each of 15 years. The hazard is summed overtime and exponentiated to calculate a risk for the disease at each year. The model is trained on a binary cross entropy loss at each year. 

\subsection{Analyses}

We consider the labeled data to be all data that we have labels for up through censoring. Since there are no ground truth treatment effects, we compare using our before and after approach to a linear regression across breast cancer patients treated with anthracyclines or trastuzumab to controls with other treatments. For the control cohort, we take the data anchored at the time of closest ECG after cancer diagnosis. The treated cohort is anchored at the time of the post treatment ECG. Our regressions account for the following variables as of the time of the measured ECG: age, sex, diabetes status, hypertension status, mean BMI, mean LDL, mean HDL, race and the following other cancer treatments: surgery, immune checkpoint inhibitor therapy, targeted therapy, monoclonal antibodies, other chemotherapy, hormone therapy, other immunotherapy, and radiation.

\subsection*{Data Availability}

Due to data use agreements and privacy, the real ECG data used in this study cannot be made available.

\subsection*{Code Availability}

Code for the study is available at \href{https://github.com/AlaaLab/before-and-after-surrogates-simulations}{AlaaLab/before-and-after-surrogates-simulations}.

\subsection*{Contributions}

FD conceived of the study and experiments. GT and JB led the real data collection. FD, AN, and AA led statistical methods. FD executed experiments and wrote the first draft. All authors contributed to writing.

% \printbibliography
% \newpage
\bibliography{biblio}
\bibliographystyle{plainnat}

\appendix
\newpage

\section{Connection to Classical Surrogate Inference, \cite{athey2025surrogate}, and \cite{ji2025predictionssurrogatesrevisitingsurrogate}}\label{related}

Our statistical surrogacy assumption (Assumption 6) is a longitudinal analogue of the Prentice surrogate, a decades old statistical concept from \cite{prentice1989surrogate} considered in the use of traditional biomarkers. We use $D$ instead of $A$ due to our unique treated vs untreated dataset set up, however, the concept is equivalent. Let $A, X, Y$ be random variables representing a treatment, a measurement, and an outcome, respectively. \\

\vspace{-.1in}
\noindent
\textbf{Definition.} (Prentice Surrogate)  A variable $X$ is a \textbf{\textit{surrogate}} for $Y$ with respect to a treatment $A$ if the null hypothesis of no relationship between $X$ and $A$ is also a valid test of the same hypothesis between $A$ and $Y$.\\

\vspace{-.1in}
\noindent
The definition is typically stated practically by studying $X, Y, A$ which satisfy the following criteria:

\begin{itemize}
    \item $A$ has an effect on $Y$,
    \item $A$ has an effect on $X$,
    \item $X$ is strongly correlated with $Y$, \textit{and}
    \item $X$ captures all of the effect of $A$ on $Y$, i.e. $Y \perp A \mid X$.
\end{itemize} 

The Prentice surrogate uses a one-dimensional $X$ to estimate the direction of the causal effect. Our AI surrogates use a regression of $Y$ on high-dimensional $X$ as the surrogate. This difference is the reason that using AI based surrogates under an analogous assumption leads to identification of the magnitude of treatment effects. If we already know $X$ to be a surrogate, then the AI model, as a regression, has nicer guarantees than the Prentice surrogate. Prentice surrogates do not eliminate the surrogate paradox without additional guarantees on monotonicity. 

Prior to our work, \cite{athey2025surrogate} studied an analogue of AI based surrogates for case and control settings. These cases and controls need not be randomized but do need to be unconfounded given a set of covariates. In their work, the objects of study are called \textit{surrogate indices} and their contribution includes novelly estimating from high-dimensional surrogates $X$. The authors train a model regressing $Y$ on $X$ unaware of time and under the analogous assumptions of Assumption 5 (comparability) and Assumption 6 (surrogacy), identify causal effects in unconfounded settings between treated and untreated groups at the same time point. Mathematically, their set up is as follows. We are given two datasets. First, we have
\begin{align}
    \mathcal{D}_{train}= \{(X, Y)\}_{i=1}^N
\end{align} for $N$ patients with labels but unknown treatment status. This is a cross sectional dataset that serves the same purpose as our retrospective longitudinal dataset. Second, we have
\begin{align}
    \mathcal{D}_{obs}= \{(X, A)\}_{i=1}^n
\end{align} for $n$ patients without labels both treated and untreated. The authors of \cite{athey2025surrogate} train the AI model (surrogate index) in $\mathcal{D}_{train}$ to approximate
\begin{align}
    \mu(X) = \mathbf{E}_{\mathcal{D}_{train}}[Y|X]
\end{align} and estimate average causal effects as
\begin{align}
    \widehat{ATE}(x) = \mathbf{E}_{\mathcal{D}_{obs}}[\widehat{\mu}(X) | A=1] - \mathbf{E}_{\mathcal{D}_{obs}}[\widehat{\mu}(X) | A=0].\footnote{We simplify that the authors include that some of the covariates can be the surrogate and other can be controls to satisfy unconfoundedness by using a single $X$.}
\end{align}

We simplify here by ignoring the inclusion of non-surrogacy covariates also included in \cite{athey2025surrogate}. We also simplify that \cite{athey2025surrogate} proposes several additional related estimators. A summary of the assumptions for effect identification under this framework are included in Table \ref{tab:athey}. The assumptions of \cite{athey2025surrogate} allow for causal inference with surrogates in clinical trials or other settings where there are sufficient covariate data to claim conditionally unconfounded treated and control groups. We add the ability to estimate heterogeneous effects using individuals as their own controls and relax the need for unconfoundedness.

Recent work in \cite{ji2025predictionssurrogatesrevisitingsurrogate} prior to ours also uses predictions as surrogates and proposes PPI with a recalibration. They begin with a general Z-estimation framework, i.e. they write
\begin{align}\label{z-estimation}
\mathbb{E}[\psi_{\theta}(X, Y)] = 0.
\end{align} ATE estimation can be written this way using, for example, outcome regression (G-computation) 
\begin{align}
    \psi_{\theta}(X, A, Y) = \mu_{A=1}(X) - \mu_{A=0}(X) - \tau
\end{align} or augmented inverse propensity weighting (AIPW)
\begin{align}
    \psi_{\theta}(X, A, Y) = \frac{A(Y- \mu_{A=1}(X))}{\pi(X)} - \frac{(1-A)(Y- \mu_{A=0}(X))}{1-\pi(X)} + \mu_{A=1}(X) - \mu_{A=0}(X) - \tau
\end{align} with ATE $\tau$, propensity model $\pi$. We can also write any M-estimation problem, as a Z-estimation problem, i.e. given $\theta^* =\mbox{argmin}_{\theta} \mathbb{E}[\ell_{\theta}(X, Y)]$, taking $\psi_{\theta} = \nabla \ell_{\theta}$ so that 
\begin{align}
    \mathbb{E}[\nabla \ell_{\theta}(X, Y)] = 0.
\end{align}
A surrogate $\hat{Y}$, unless perfectly correlated with $Y$ or otherwise proved to identify the treatment effect (i.e. in \cite{athey2025surrogate} or ours), cannot be used to estimate $\theta$ directly and will have $\mathbb{E}[\psi_{\theta^*}(X, \hat{Y})] \neq 0$ for optimal $\theta^*$ solving Equation \ref{z-estimation} \citep{robins1994estimation}. The solution in classical surrogate literature \citep{robins1994estimation} is to use an augmented inverse probability weighted estimator, where for probability of label missingness $p=P(D=1|Y,\hat{Y},X)$, we write
\begin{align}
    \sum_{i=1}^{n+N}\frac{D_i}{p} \psi_{\theta}(X_i, Y_i) - \frac{D_i-p}{p} s_{\theta}(X_i, Y_i)=0.
\end{align}

\cite{robins1994estimation} shows the optimal choice of $s$ is $s(X,\hat{Y}):=\mathbb{E}[ \psi_{\theta}(X, Y) | X, \hat{Y}]$. The idea of \cite{ji2025predictionssurrogatesrevisitingsurrogate} is apply this framework to the PPI estimator \citep{angelopoulos2023ppi++} to improve calibration.  As an M-estimation problem, PPI is written \begin{align}\label{base_ppi_m_est}
  \theta^{PPI} = \mbox{argmin}_{\theta} \frac{1}{n} \sum_{i=1}^n\ell_{\theta}(X, Y)  - \left( \frac{1}{n} \sum_{i=1}^n g_{\theta}(X, \hat{Y}) - \frac{1}{N} \sum_{i=n+1}^{n+N} g_{\theta}(X, \hat{Y}) \right), 
\end{align} for some $g$ (i.e. $g=\ell$ in the base case). \cite{ji2025predictionssurrogatesrevisitingsurrogate} uses the surrogate score $s(X,\hat{Y})$. The score uses the Z-estimation version of the PPI problem, so they take $\psi_{\theta} = \nabla \ell_{\theta}$ and estimate \begin{align}
    \hat{s}(X,\hat{Y}) \approx \mathbb{E}[\nabla \ell_{\theta}(X,Y ) \mid X, \hat{Y}].
\end{align} Then they define
\begin{align}
    \hat{g}_{\theta}(X,\hat{Y}) \propto \hat{M}\cdot \hat{s}(X,\hat{Y}).
\end{align} The approach in \cite{ji2025predictionssurrogatesrevisitingsurrogate} requires additionally estimating $\hat{s}$ and a tuning weight matrix $\hat{M} = \mbox{Cov}(\nabla \ell_{\theta} (X, Y) , \hat{s}(X, \hat{Y}))\cdot \mbox{Cov}(\hat{s}(X, \hat{Y}))^{-1}$ with labeled data. Omitting precise details, the authors of \cite{ji2025predictionssurrogatesrevisitingsurrogate} use cross validation for their estimator and ultimately obtain an result that is always statistically better than using only the labeled data (an improvement over simple PPI). This method (RePPI) reduces issues in using poor predictions as surrogates by recalibrating. RePPI might be used to improve inference with before-and-after surrogates as well.

Notice that all standard identification requirements of any estimand $\psi_{\theta}$ in the new estimator proposed by \cite{ji2025predictionssurrogatesrevisitingsurrogate} are still required. Our estimator, by contrast proposes a new framework and identification for treatment effect estimation that may be either used without correction (Proposition 2) or to which the estimator \cite{ji2025predictionssurrogatesrevisitingsurrogate} might be extended. This (or PPI++ \cite{angelopoulos2023ppi++}, etc.) may further improve inference in settings with poor surrogates.

There are other works using AI models to estimate treatment effects in settings where we have access to the treatment at training \citep{PPCI, PPgCI}. These methods have utility when we have data on the treatment effect and seek to extend it for faster inference with an AI surrogate but cannot be used prior to having large quantities of data on treated outcomes, a benefit of our approach.

%\textcolor{red}{\textbf{Figure?}. Athey framework as comparison}
\begin{table}[h!]
    \centering
    \begin{tabular}{ccc}\toprule
        & Assumption&  Testable?  \\\midrule
       \textbf{Standard} &  Consistency& 
     No \\ 
& Positivity& Yes\\
& Unconfoundedness & No  \\ 
\textbf{AI Surrogacy}&Surrogacy& No \\ 
& Comparability& No \\
 \bottomrule\end{tabular}
    \caption{The assumptions for inference with case control surrogates are similar to ours but require unconfoundedness (as identified in \cite{athey2025surrogate}).}
    \label{tab:athey}
\end{table}

\section{Identification Proof}\label{app:proofs}

We briefly re-introduce our mathematical set up. Recall that we have time $T$ with two observation points for the treated patients $t\in \{t^{pre},t^{post}\}$ and a binary treatment $A^t \in \{0,1\}$ applied between the two measurements, i.e. $A^{t^{post}}$=1 and $A^{t^{pre}}$=0. We observe for each patient measurement data $X^t_i \in \mathcal{X}$. Treated patients have measurements observed at both $t^{pre}$ and $t^{post}$. Our outcome of interest is $Y^t \in \mathbb{R}$ measured at some time $t^{end}$.

Following the potential outcomes framework \citep{rubin1974estimating, imbens2015causal}, each subject has potential outcomes $Y^{t^{end}}(0)$ for no treatment by $t^{post}$ and $Y^{t^{end}}(1)$ for treatment occurring by $t^{post}$. We have potential measurements as well, i.e. $X^{t^{pre}}(0), X^{t^{post}}(0)$ for no treatment by $t^{post}$ and $X^{t^{pre}}(1), X^{t^{post}}(1)$ for treatment occurring by $t^{post}$. Since $A^{t^{post}}$ cannot affect $X^{t^{pre}}$, $X^{t^{pre}}(0) = X^{t^{pre}}(1)$. 

In our setup, we have access to an intervention-free training dataset $$\mathcal{D}_{train} = \{(X_i^{t_i^1},X_i^{t^2_i},\ldots, Y^{t^1_i}_i,Y^{t^2_i}_i, \ldots)\}_{i=1}^N,$$ in addition to an (unlabeled) observational dataset $$\mathcal{D}_{obs}=\{(X^{t_i^{pre}}_i, X^{t_i^{post}}_i, A^{t_i}_i)\}_{i=1}^n.$$ We let $D = 1$ denote a sample from $\mathcal{D}_{\text{obs}}$ and $D = 0$ a sample from $\mathcal{D}_{\text{train}}$. 

Given $\mathcal{D}_{train}$ and $\mathcal{D}_{obs}$, our goal is to estimate the average or conditional average treatment effect (ATT or CATT) on the treated. Recall,
\begin{align}
\mbox{CATT}(x)= \mathbb{E}\left[Y^{t^{end}}(1) - Y^{t^{end}}(0) \mid X^{t^{pre}}=x, A^{t^{post}}=1, D=1 \right].
\end{align} 
and 
\begin{align}
\mbox{ATT} = \mathbb{E}[\mbox{CATT}(x) \mid A^{t^{post}}=1, D=1].
\end{align}
However, because outcomes are never observed in $\mathcal{D}_{\text{obs}}$, the CATT nor ATT cannot be estimated directly via standard covariate adjustment with a control group. To identify CATT and ATT from $\mathcal{D}_{\text{train}}$ and $\mathcal{D}_{\text{obs}}$, we make a series of assumptions. These are listed in the main text but restated here. In the main text we focus on ATT, as we can only obtain valid inference by averaging. We include the CATT identification result here as well. We begin with two standard conditions for causal inference in the potential outcomes framework \citep{imbens2015causal, rosenbaum1983central}:\\

\vspace{-.1in}
\noindent \textbf{Assumption 1} (Consistency). For $A^{t^{post}}=a$ and $t\in\{t^{pre}, t^{post}\}$, $Y^t=Y^t(a)$ and $X^t=X^t(a)$.\\

\vspace{-.1in}
\noindent \textbf{Assumption 2} (Positivity). For $t \in \{t^{pre}, t^{post}\}$ and $x \in \mathcal{D}_{obs}$, $P(D=0 \mid X^t =x) > 0.$\\

\vspace{-.1in}
\noindent \textbf{Assumption 3} (Covariate Markov Sufficiency). For $d \in \{0, 1\}$ and $t'' > t' > t$, $Y^{t''} \perp X^{t} \mid X^{t'}, D=d$.
\\

\vspace{-.1in}
\noindent \textbf{Assumption 4} (Counterfactual Baseline Sufficiency) $Y^{t^{end}}(0) \perp X^{t^{post}}(0) \mid X^{t^{pre}}(0)$.\\

\vspace{-.1in}
\noindent
\textbf{Assumption 5} (Comparability). For $t' > t$, $Y^{t'}(0) \perp D \mid X^{t}(0)$.\\

\vspace{-.1in}
\noindent
\textbf{Assumption 6} (Statistical Surrogacy). For any $t$ with $t'> t\geq t^{post}$, $Y^{t'} \perp D \mid X^{t}$.\\

\noindent \textbf{Proposition 1 and 4.} (\textit{Identification of ATT and CATT})
Let $\mu(x, t)$ be a function (also defined in Equation \ref{model}) representing the expected value of the future outcome $t$ years from the measurement of $x$ conditional on the measurement $x$ in the training distribution:
\begin{align}\label{appendix_mu_def}
    \mu(X^{t^0}, t^1 - t^0) = \mathbb{E}\left[Y^{t^1} \mid X^{t^0},  D=0\right].
\end{align} Then, under assumptions 1-6, we have:
\[\mbox{CATT}(x)=\mathbb{E}[\mu(X^{t^{post}},  {t^{end}}-{t^{post}}) -
    \mu(x,  {t^{end}}-{t^{pre}})  | X^{t^{pre}}=x, A^{t^{post}}=1, D=1]\]
and 
\[\mbox{ATT}=\mathbb{E}[\mu(X^{t^{post}},  {t^{end}}-{t^{post}}) -
    \mu(X^{t^{pre}},  {t^{end}}-{t^{pre}})  | A^{t^{post}}=1, D=1].\] \\

\noindent \textit{Proof.}
Since $A^{t^{post}}=1$ for all subjects in $\mathcal{D}_{obs}$, conditioning on $D=1$ is equivalent to conditioning on $\{D=1, A^{t^{post}}=1\}$. Thus, {CATT}(x) rearranging the definition can be written as follows:
\[\mbox{CATT}(x) = \mathbb{E}[Y^{t^{end}}(1) \mid X^{t^{pre}}=x, D=1] - \mathbb{E}[Y^{t^{end}}(0) \mid X^{t^{pre}}=x, D=1].\]
We handle the two potential outcome terms above separately.\\

\vspace{-.1in}
\noindent \textit{\textbf{Identification of $\boldsymbol{\mathbb{E}[Y^{t^{end}}(1) \mid X^{t^{pre}}=x, D=1]}$:}
}

 By the tower property, conditioning on $X^{t^{post}}(1)$, we have:
\begin{align}
\mathbb{E}\left[Y^{t^{end}}(1) \mid X^{t^{pre}}=x, D=1 \right] = \mathbb{E}\left[\mathbb{E}\left[Y^{t^{end}}(1) \mid X^{t^{post}}(1), X^{t^{pre}}=x, D=1\right] \mid X^{t^{pre}}=x, D=1\right].
\label{eq::1::prop1}
\end{align}
By consistency (Assumption 1), followed by the fact that all in $\mathcal{D}_{obs}$ are treated, we write
\begin{align}
\mathbb{E}\left[Y^{t^{end}}(1) \mid X^{t^{post}}(1), X^{t^{pre}}=x, D=1\right] &= \mathbb{E}\left[Y^{t^{end}} \mid X^{t^{post}}, X^{t^{pre}}=x, A^{t^{post}}=1, D=1\right] \label{eq::part_1:25}\\
&= \mathbb{E}\left[Y^{t^{end}} \mid X^{t^{post}}, X^{t^{pre}}=x, D=1\right]
\end{align}
By Markov covariate sufficiency (Assumption 3), $Y^{t^{end}} \perp X^{t^{pre}} \mid X^{t^{post}}$, hence we have
\begin{align}
\mathbb{E}\left[Y^{t^{end}} \mid X^{t^{post}}, X^{t^{pre}}=x, D=1\right] = \mathbb{E}\left[Y^{t^{end}} \mid X^{t^{post}}, D=1\right].
\label{eq::2::prop1}
\end{align}

\noindent Using  surrogacy (Assumption 6), we can write (\ref{eq::2::prop1}) as 
\begin{align}
\mathbb{E}\left[Y^{t^{end}} \mid X^{t^{post}}, D=1\right] &= \mathbb{E}\left[Y^{t^{end}} \mid X^{t^{post}}, D=0\right].
\label{eq::4::prop1}
\end{align}
Combining (\ref{eq::part_1:25}), (\ref{eq::2::prop1}), and (\ref{eq::4::prop1}), we get
\begin{align}
\mathbb{E}\left[Y^{t^{end}}(1) \mid X^{t^{post}}(1), X^{t^{pre}}=x, D=1\right] &= \mathbb{E}\left[Y^{t^{end}} \mid X^{t^{post}}, D=0\right] \\ &= \mu(X^{t^{post}}, t^{end} - {t^{post}}).
\label{eq::5::prop1}
\end{align}
By plugging (\ref{eq::5::prop1}) back into (\ref{eq::1::prop1}), we recover the identification result for the first potential outcome:
\begin{align}
\mathbb{E}\left[Y^{t^{end}}(1) \mid X^{t^{pre}}=x, A^{t^{post}}=1, D=1 \right] = \mathbb{E}\left[\mu(X^{t^{post}},t^{end}-t^{post})) \mid X^{t^{pre}}=x, D=1\right].
\tag{$\star$}
\label{eq::6::prop1}
\end{align}

\noindent \textit{\textbf{Identification of $\boldsymbol{\mathbb{E}[Y^{t^{end}}(0) \mid X^{t^{pre}}=x, D=1]}$:}}

This is the counterfactual quantity since all subjects in $\mathcal{D}_{obs}$ have $A^{t^{post}}=1$, hence $Y^{t^{end}}(0)$ is never observed. Since the treatment has not happened yet at time $t={t^{pre}}$, $X^{t^{pre}}(1) = X^{t^{pre}}(0)$, so rewriting then using consistency (Assumption 1) we have
\begin{align}
\mathbb{E}[\mu(X^{t^{pre}}, {t^{end}} - t^{pre}) \mid X^{t^{pre}}=x, D=1] &= \mu(x, {t^{end}} - t^{pre}) \nonumber \\& = \mathbb{E}\left[ Y^{t^{end}} \mid X^{t^{pre}}=x, D=0 \right] \\& = \mathbb{E}\left[ Y^{t^{end}} \mid X^{t^{pre}}(0)=x, D=0 \right] 
\label{eq::7::prop1}
\end{align}

Using by line counterfactual baseline sufficiency (Assumption 4), definition of potential outcomes, counterfactual baseline sufficiency (Assumption 4) then comparability (Assumption 5) in sequence, we can write
\begin{align}
    \mathbb{E}\left[ Y^{t^{end}} \mid X^{t^{pre}}(0)=x, D=0 \right] 
    &= \mathbb{E}\left[ Y^{t^{end}} \mid X^{t^{post}}(0), X^{t^{pre}}(0)=x, D=0 \right] \\ &= \mathbb{E}\left[ Y^{t^{end}}(0) \mid X^{t^{post}}(0), X^{t^{pre}}(0)=x, D=0 \right] \\&= \mathbb{E}\left[Y^{t^{end}}(0) \mid X^{t^{pre}}(0)=x, D=0 \right]\\ & = \mathbb{E}\left[ Y^{t^{end}}(0) \mid X^{t^{pre}}(0)=x, D=1\right].\tag{$\star\star$} \label{eq::9::prop1}
\end{align}

Subtracting (\ref{eq::9::prop1}) from (\ref{eq::6::prop1}), we can write the CATT as:
\[\mbox{CATT}(x) = \mathbb{E}\left[\mu(X^{t^{post}}, {t^{end}} - t^{post}) - \mu(x, {t^{end}} - t^{pre})\mid X^{t^{pre}}=x, D=1\right].\]

As a direct corollary, we have identification of ATT as $\mathbb{E}[\mbox{CATT}(x) \mid D=1]$. $\square$ 

\section{Covariate Shift Estimators}\label{app:estimators}

Notice that in addition to Assumptions 1-6, in \textbf{Proposition 3}, we assumed that the pre-treatment data came from the same distribution as the training data. When we cannot assume that the pre-treatment data in $\mathcal{D}_{obs}$, i.e. $(X^{t^{pre}}, Y^{t^{end}}(0))$, comes from the same distribution as $\mathcal{D}_{train}$, we cannot use \textbf{Proposition 3} and Equation \ref{ppi_estimator_eqn} directly. This may occur, for example, when the treated population in $\mathcal{D}_{obs}$ is especially ill compared to a general population. We can, however, leverage a prediction powered inference \citep{ppi} correction now additionally with a covariate shift. Assumptions 5 and 6 allow us to view changes in the distributions between each of $(X^{t^{pre}}, Y^{t^{end}}(0))\sim P_{obs, pre}$, $(X^{t^{post}}, Y^{t^{end}}(1))\sim P_{obs, post}$ and some data $(X^{t}, Y^{t'})\sim P_{train}$ as a covariate shift. This is since, in Assumptions 5 and 6, we fix the relationship between the covariates $X^t$ and the outcomes $Y^t$. 

The first estimator below (Equation \ref{ATT_tilde}) does not require labels in the treated population, only labels on a sample held out from the training data (by assuming Assumptions 1-6). The second application of covariate shift relaxes surrogacy (relaxes Assumption 6 and assumes Assumptions 1-5), uses a sample of labeled treated data, and a covariate shift to correct the pre-treatment error (Equation \ref{ppi_estimator_eqn_w_cov}). We introduce the additional estimators and results for valid inference with them here. Proofs for inference results follow in the next section, Appendix \ref{app:estimators_proofs}. There are many indices in writing out the estimators. We maintain the convention that the index $i$ is for individuals in $\mathcal{D}_{obs}$ and $j$ for individuals in $\mathcal{D}_{train}$. 

\subsection{Estimators with covariate shift assuming surrogacy}

Select a collection of points from the training distribution with the same fixed time interval (later notated as $k$) as the difference between $t^{end}$ and $t^{pre}$: $(X^{t^{0}}, Y^{t^k}) \sim  P_{train}$. Write the density ratio $w(x)$ between the distribution of pre- treatment observed data $(X^{t^{pre}}, Y^{t^{end}}(0))\sim P_{obs, pre}$ and this fixed interval training data:
\begin{align}
    w_{pre}(x):=\frac{dP(X^t=x|D=1,t=t^{pre})}{dP(X^{t^0}=x|D=0)}
\end{align} and analogously that for the post treatment data $(X^{t^{post}}, Y^{t^{end}}(1))\sim P_{obs, post}$ and training data $(X^{t^{0}}, Y^{t^{k'}}) \sim  P_{train}$ with fixed time interval $t^{end} - t^{post}$ (denoted as $k'$): \begin{align}
    w_{post}(x):=\frac{dP(X^t=x|D=1,t=t^{post})}{dP(X^{t^0}=x|D=0)}.
\end{align} 

\noindent Let $\hat{\mu}$ be a (not necessarily unbiased) estimator of $\mu$. Write for model $\widehat{\mu}$ approximating $\mu$ (Equation \ref{appendix_mu_def}), the ATT estimator of \begin{align}
   \widehat{\text{ATT}} =\frac{1}{n}\sum_{i=1}^n\left(\widehat{
   \mu} (X_i^{t_i^{post}},{t_i^{end}} - {t_i^{post}} ) -
   \widehat{
   \mu} (X_i^{t_i^{pre}},{t_i^{end}} - {t_i^{pre}} ) \right).
\end{align} Write, by linearity, the difference between true ATT and expected value of the $\widehat{\mbox{ATT}}(x)$ as
\begin{align}\label{bias_terms}
\mbox{ATT} - \mathbb{E}[\widehat{\mbox{ATT}}]
&=
\underbrace{\mathbb{E}_{{obs, post}}\left[
Y_i^{t_i^{end}}(1)
-
\widehat{\mu}(X_i^{t_i^{post}}, t_i^{end}-t_i^{post})
\mid A^{t^{post}}=1, D=1
\right]}_{\text{How wrong we are on }Y^{t^{end}}(1)}  \nonumber
\\
&\quad -
\underbrace{\mathbb{E}_{{obs, pre}}\left[
Y_i^{t_i^{end}}(0)
-
\widehat{\mu}(X_i^{t_i^{pre}}, t_i^{end}-t^{pre})
\mid A^{t^{post}}=1, D=1
\right].}_{\text{How wrong we are on }Y^{t^{end}}(0)}
\end{align} We emphasize the distribution over which the expectation is taken using the subscripts. We seek in this section to rewrite these bias terms using terms we can instead estimate with data. Differently from the maintext results, we now allow for distribution shifts between the distributions that $\mathcal{D}_{obs}$ and  $\mathcal{D}_{train}$ come from.\\

\noindent \textbf{Proposition 4.} \textit{(Covariate Shift Identification)} Assume Assumptions 1-6. Then we have,
\begin{align}\label{cov_shift_no_labels}
    \mbox{ATT} = \mathbb{E}[\widehat{\mbox{ATT}}] - \mathbb{E}_{train}[w_{pre}(X^{t^0}) \cdot (Y^{t^k} - \hat{Y}^{t^k})] + \mathbb{E}_{train}[w_{post}(X^{t^0}) \cdot (Y^{t^{k'}} - \hat{Y}^{t^{k'}})].
\end{align} \\

\noindent \textit{Proof.}
    It suffices to rewrite each term of Equation \ref{cov_shift_no_labels} as equal to a term in Equation \ref{bias_terms}. First, we rewrite then we re-weight to obtain a change of measure from $P_{train}$ to $P_{obs, pre}$,
    \begin{align}
        \mathbb{E}_{train}[w_{pre}(X^{t^0}) \cdot (Y^{t^k} - \hat{Y}^{t^k})]  &= \mathbb{E}_{train}[ w_{pre}(X^{t^0}) \cdot(Y^{t^k} - \hat{\mu}(X^{t^0}, t^k - t^0)]\\
        &= \mathbb{E}_{obs, pre}[ (Y^{t^k} - \hat{\mu}(X^{t^0}, t^k - t^0)]
    \end{align}
    and to obtain a change of measure from $P_{train}$ to $P_{obs, post}$,
    \begin{align}
        \mathbb{E}_{train}[w_{pre}(X^{t^0}) \cdot (Y^{t^k} - \hat{Y}^{t^k})]   &= \mathbb{E}_{train}[ w_{post}(X^{t^0}) \cdot(Y^{t^{k'}} - \hat{\mu}(X^{t^0}, t^{k'} - t^0)]\\
        &= \mathbb{E}_{obs, post}[ (Y^{t^{k'}} - \hat{\mu}(X^{t^0}, t^{k'} - t^0))].
    \end{align} Then as $k, k'$ match the intervals from the pre and post distributions, respectively, and the training data are intervention-free, these are exactly the desired intervention-free expectations. Data with time gap $k$ in the observed pre-treatment distribution are distributed like $(X^{t^{pre}}, Y^{t^{end}}(0)) \sim P_{obs,pre}$. Data with time gap $k'$ in the observed post treatment distribution are distributed like $ (X^{t^{post}}, Y^{t^{end}}(1)) \sim P_{obs,post}$.  $\square$ \\

\noindent \textbf{Proposition 5.} \textit{(Covariate Shift Estimator and Inference)}
Assume Assumptions 1-6. Calculate for each $i$ individual of the $n$ total in $\mathcal{D}_{obs}$:
\begin{align}
    \widehat{Y}_i^{t_i^{end}}(0)  = \widehat{\mu}(X_i^{t_i^{pre}}, t_i^{end}-t_i^{pre}), \,\,
    \widehat{Y}_i^{t_i^{end}}(1)  = \widehat{\mu}(X_i^{t_i^{post}}, t_i^{end}-t_i^{post}).
\end{align} Divide the $N$ labeled, randomly ordered individuals sampled from held out data in $\mathcal{D}_{train}$ into two groups of size $N/2$ (should $N$ be odd, take $N/2$ as the floor). For one group, we will compare predictions and labels using the time gap $k = t^{end}-t^{pre}$ (predict from $X^{t^{0}}$ the label $Y^{t^{k}}$). For the second group, we will compare predictions and labels using the time gap $k' = t^{end}-t^{post}$ (predict from $X^{t^{0}}$ and $Y^{t^{k'}}$). Thus, we take
\begin{align}
    \widehat{Y}_j^{t_j^k}  = \widehat{\mu}(X_j^{t_j^0}, t_j^k-t_j^0), \, \, \widehat{Y}_j^{t_j^{k'}}  = \widehat{\mu}(X_j^{t_j^0}, t_j^{k'}-t_j^0).
\end{align} 
Plug in an estimator $\hat{w}$ for each density ratio $w$ (as done in \cite{shimodaira2000improving}). Define the estimator $\widetilde{ATT}^{PPI}$ to be
    \begin{align}\label{ATT_tilde}
        \frac{1}{n}\sum_{i=1}^{n} (\widehat{Y}_i^{t_i^{end}}(1) - \widehat{Y}_i^{t_i^{end}}(0)) -  \frac{1}{N/2}\sum_{j=1}^{N/2} \hat{w}_{pre}(X_j^{t_j^0})\cdot (Y_j^{t_j^k} - \widehat{Y}_j^{t_j^k}) +  \frac{1}{N/2}\sum_{j=N/2+1}^{N} \hat{w}_{post}(X_j^{t_j^0})\cdot (Y_j^{t_j^{k'}} - \widehat{Y}_j^{t_j^{k'}}) 
    \end{align} 
    with sample variances \begin{align*}
        \widehat{\sigma}_{ATT}^2 &=\frac{1}{n-1}\sum_{i=1}^{n}({\Delta_{ATT}}_i - {\bar{\Delta}_{ATT}})^2 \\
        \widehat{\sigma}_{post}^2 &=\frac{1}{N/2-1}\sum_{j=1}^{N/2}({\Delta_{post}}_j - {\bar{\Delta}_{post}})^2 \\
        \widehat{\sigma}_{pre}^2 &=\frac{1}{N/2-1}\sum_{j=N/2+1}^{N}({\Delta_{pre}}_j - {\bar{\Delta}_{pre}})^2
    \end{align*} 
    for ${\Delta_{ATT}}_i =\widehat{Y}_i^{t_i^{end}}(1) - \widehat{Y}_j^{t_j^{end}}(0)$ with mean $\bar{\Delta}_{ATT}$, ${\Delta_{post}}_j =\hat{w}_{post}(X_j^{t_j^0})\cdot (Y_j^{t_j^{k'}} - \widehat{Y}_j^{t_j^{k'}})$ with mean $\bar{\Delta}_{post}$, and ${\Delta_{pre}}_j =\hat{w}_{pre}(X_j^{t_j^0})\cdot (Y_j^{t_j^k} - \widehat{Y}_j^{t_j^k})$ with mean $\bar{\Delta}_{pre}$. Then with $z_{1-\alpha/2}$ denoting the $1-\alpha/2$ quantile of the standard normal distribution, the prediction powered confidence interval created \begin{align}
        \widetilde{ATT}^{PPI} \pm z_{1-\alpha/2} \sqrt{\frac{\widehat{\sigma}_{ATT}^2}{n} + \frac{\widehat{\sigma}_{post}^2}{N/2} + \frac{\widehat{\sigma}_{pre}^2}{N/2}}
    \end{align} has asymptotic coverage of ATT at level $1-\alpha$ assuming $||\hat{w} - w||_{L_2(P_{\mathcal{D}_{train}})} = o_p(N^{-1/2})$ for each of $w_{pre}, w_{post}$ and finite true variance and consistent variance estimators.\footnote{We use the $o_p$ notation as is standard for the convergence to zero in probability. This is of course an assumption on how good the covariate shift model is. Bias in this model is not corrected for, which is a limitation.}

See Appendix \ref{cov_inf_proof} for the proof.

\subsection{Estimators with covariate shift in the absence of surrogacy}

In the absence of surrogacy (Assumption 6), we can use a small labeled set of treated individuals to adjust for the bias in how wrong we are in predicting $Y^{t^{end}}(1)$, which is now due additionally to lack of surrogacy. We now present results that leverage prediction powered inference and covariate shift when surrogacy can fail. Since we use the real labels for $Y^{t^{end}}(1)$, we do not need to covariate shift the error on the post treatment data. \\

\noindent \textbf{Proposition 7.} \textit{(Covariate Shift Identification without Surrogacy)} Assume Assumptions 1-5. Then we have,
\begin{align}
    \mbox{ATT} = \mathbb{E}[\widehat{\mbox{ATT}}] - \mathbb{E}_{train}[w_{pre}(X^{t^0}) \cdot (Y^{t^k} - \hat{Y}^{t^k})] + \mathbb{E}[Y^{t^{end}}(1) - \hat{\mu}(X^{t^{post}}, t^{end}-t^{post})].
\end{align}

\textit{Proof.}
    We can write $\mbox{ATT}-\mathbb{E}[\widehat{\mbox{ATT}}]$ as the same difference in Equation \ref{bias_terms} above. Thus, it suffices to show the pre treatment covariate shift matches. As no assumptions on these data have changed, this follows the same as in Proposition 4. $\square$ \\

\noindent \textbf{Proposition 8.} \textit{(Covariate Shift Estimator without Surrogacy)} 
    Assume Assumptions 1-5. Suppose we have labels $Y^{t^{end}}(1)$ for a size $n_L$ subset of $\mathcal{D}_{obs}$. Calculate for each individual in $\mathcal{D}_{obs}$:
\begin{align}
    \widehat{Y}_i^{t_i^{end}}(0)  = \widehat{\mu}(X_i^{t_i^{pre}}, t_i^{end}-t_i^{pre}), \,\,
    \widehat{Y}_i^{t_i^{end}}(1)  = \widehat{\mu}(X_i^{t_i^{post}}, t_i^{end}-t_i^{post}).
\end{align} Calculate for the individuals in $\mathcal{D}_{train}$ with the same time gap $k$ between $X^{t^{pre}}$ and $Y^{t^{end}}$ as between $t^{pre}$ and $t^{end}$:
\begin{align}
    \widehat{Y}_i^{t_i^k}  = \widehat{\mu}(X_i^{t_i^0}, t_i^k-t_i^0).
\end{align} Write the density ratio $w(x)$ between the distribution of pre treatment observed data $(X^{t^{pre}}, Y^{t^{end}})\sim \mathcal{D}_{obs, pre}$ and training data $(X^{t^{0}}, Y^{t^k}) \sim \mathcal{D}_{train}$ as
\begin{align}
w(x):=\frac{dP(X^t=x|D=1,t=t^{pre})}{dP(X^{t^0}=x|D=0)}.
\end{align}
Plug in an estimator $\hat{w}$ of $w$, and define the estimator $\widetilde{ATT}^{PPI-L}$\footnote{$L$ for treated \textbf{L}abels.} to be \begin{align}\label{ppi_estimator_eqn_w_cov}
        \frac{1}{n-n_L}\sum_{i=n_L+1}^{n} (\widehat{Y}_i^{t_i^{end}}(1) - \widehat{Y}_i^{t_i^{end}}(0)) +  \frac{1}{n_L}\sum_{i=1}^{n_L} (Y^{t_i^{end}}_i(1) - \widehat{Y}_i^{t_i^{end}}(1)) -  \frac{1}{N}\sum_{j=1}^N \hat{w}(X^{t_j^0})\cdot (Y_j^{t_j^k} - \widehat{Y}_j^{t_j^k}) 
    \end{align} 
    with sample variances \begin{align*}
        \widehat{\sigma}_{ATT}^2 &=\frac{1}{n-n_L-1}\sum_{i=n-n_L+1}^{n}({\Delta_{ATT}}_i - {\bar{\Delta}_{ATT}})^2 \\
        \widehat{\sigma}_{post}^2 &=\frac{1}{n_L-1}\sum_{i=1}^{n_L}({\Delta_{post}}_i - {\bar{\Delta}_{post}})^2 \\ \widehat{\sigma}_{control}^2 &=\frac{1}{N-1}\sum_{j=1}^N({\Delta_{control}}_j - {\bar{\Delta}_{control}})^2
    \end{align*} 
    for ${\Delta_{ATT}}_i =\widehat{Y}_i^{t_i^{end}}(1) - \widehat{Y}_i^{t_i^{end}}(0)$ with mean $\bar{\Delta}_{ATT}$, ${\Delta_{post}}_i ={Y}_i^{t_i^{end}}(1) - \widehat{Y}_i^{t_i^{end}}(1)$ with mean $\bar{\Delta}_{post}$, and ${\Delta_{control}}_j =\hat{w}(X^{t_j^0})\cdot (Y_j^{t_j^k} - \widehat{Y}_j^{t_j^k})$ with mean $\bar{\Delta}_{control}$, respectively. Then with $z_{1-\alpha/2}$ denoting the $1-\alpha/2$ quantile of the standard normal distribution, the prediction powered confidence interval created \begin{align}
        \widetilde{ATT}^{PPI-L} \pm z_{1-\alpha/2} \sqrt{\frac{\widehat{\sigma}_{ATT}^2}{n-n_L} + \frac{\widehat{\sigma}_{post}^2}{n_L} + \frac{\widehat{\sigma}_{control}^2}{N}}
    \end{align} has asymptotic coverage of ATT at level $1-\alpha$ assuming $||\hat{w} - w||_{L_2(P_{\mathcal{D}_{train}})} = o_p(N^{-1/2})$ and finite true variance and consistent variance estimators.

See Appendix \ref{cov_inf_proof} for the proof of \textbf{Proposition 8}. The result that appears in the main text is a weaker version of the above in that it assumes the distribution of pre treatment images is the same as that of the training controls. This is reasonable since the pre treatment data does not receive treatment but may not hold if treated individuals are a particular subset of the general population due to their need for treatment. 

\section{Inference Proofs}\label{app:estimators_proofs}

\subsection{Main text results}

\textbf{Proposition 2.} Assume Assumptions 1-6 and that the model bias $\mathbb{E}[Y^{t+t'} - \hat{\mu}(X^t, t')]$ is constant in $t'$ and between samples with \textit{unobserved} labels $(X^{t^{pre}}, Y^{t^{pre}+t'}(0))$ and $(X^{t^{post}}, Y^{t^{post}+t'}(1))$. Define the estimator $\widehat{ATT}$ as
    \begin{align*}
        \frac{1}{n}\sum_{i=1}^{n} (\widehat{Y}_i^{t_i^{end}}(1) - \widehat{Y}_i^{t_i^{end}}(0))
    \end{align*} 
    with sample variance $\widehat{\sigma}_{ATT}^2 = \frac{1}{n-1}\sum_i^n(\Delta_i - \bar{\Delta})^2$ for $\Delta_i =\widehat{Y}_i^{t_i^{end}}(1) - \widehat{Y}_i^{t_i^{end}}(0)$ with mean $\bar{\Delta}$. Then with $z_{1-\alpha/2}$ denoting the $1-\alpha/2$ quantile of the standard normal distribution, the interval created \begin{align}\label{estimator1}
        \widehat{ATT} \pm z_{1-\alpha/2} \sqrt{\frac{\widehat{\sigma}_{ATT}^2}{n}}
    \end{align} has asymptotic coverage of ATT at level $1-\alpha$ assuming finite and consistent true variance.  \\

\textit{Proof.}
    We begin by showing that $\mathbb{E}[\widehat{\mbox{ATT}}] =$ ATT. We do so by showing $\mbox{ATT} - \mathbb{E}[\widehat{\mbox{ATT}}] =0$. First, we write this difference in terms of $\widehat{\mu}$ as 
    \begin{align}
    = \mathbb{E}\left[
    Y_i^{t_i^{end}}(1)
    -
    \widehat{\mu}(X_i^{t_i^{post}}, t_i^{end}-t_i^{post})
    \mid A^{t^{post}}=1, D=1
    \right]
    \\
     - \mathbb{E}\left[
    Y_i^{t_i^{end}}(0)
    -
    \widehat{\mu}(X_i^{t_i^{pre}}, t_i^{end}-t_i^{pre})
    \mid A^{t^{post}}=1, D=1
    \right]
    \end{align} leveraging the linearity of expectation as above. Then, since the bias is assumed constant across samples $(X^{t^{pre}}, Y^{t^{pre}+t'}(0))$ and $(X^{t^{post}}, Y^{t^{post}+t'}(1))$ and over time $t'$, these are equal and cancel.

    The rest of the proof is standard application of the Central Limit Theorem. We assume iid terms, finite variance, and that $\hat{\sigma}_{ATT}$ is consistent estimator of the true variance. This gives that for large enough $n$,  
    \begin{align}
        \sqrt{n}(\widehat{ATT} - \mathbb{E}[\widehat{ATT}]) \Rightarrow \mathcal{N}(0, {\sigma}^2_{ATT})
    \end{align} implying that the desired interval has the desired asymptotic coverage of ATT. $\square$ \\

\textbf{Proposition 3.} Assume Assumptions 1-5. Assume additionally that $(X^{t^{pre}}, Y^{t^{end}}(0))$ can be viewed as a random sample from the same underlying distribution as equally spaced observations $(X^{t^{0}}, Y^{t^{k}})$ in $\mathcal{D}_{train}$. Imagine we have labels $Y^{t^{end}}(1)$ for a size $n_L$ subset of $\mathcal{D}_{obs}$, which has $n$ total observations. Calculate for each individual in $\mathcal{D}_{obs}$:
\begin{align*}
    \widehat{Y}_i^{t_i^{end}}(0)  = \widehat{\mu}(X_i^{t_i^{pre}}, t_i^{end}-t_i^{pre}), \,\,
    \widehat{Y}_i^{t_i^{end}}(1)  = \widehat{\mu}(X_i^{t_i^{post}}, t_i^{end}-t_i^{post}).
\end{align*} Take from the dataset $\mathcal{D}_{train}$ an independent sample not used in training of size $N$. Calculate for the individuals in $\mathcal{D}_{train}$ with the same time gap $k$ between $X^{t^{pre}}$ and $Y^{t^{end}}$ as between $t^{pre}$ and $t^{end}$:
\begin{align*}
    \widehat{Y}_j^{t_i^k}  = \widehat{\mu}(X_j^{t_i^0}, t_j^k-t_j^0).
\end{align*}
Define the estimator $\widehat{ATT}^{PPI}$ to be
    \begin{align}
        \frac{1}{n-n_L}\sum_{i=n_L+1}^{n} (\widehat{Y}_i^{t_i^{end}}(1) - \widehat{Y}_i^{t_i^{end}}(0)) +  \frac{1}{n_L}\sum_{i=1}^{n_L} (Y^{t_i^{end}}_i(1) - \widehat{Y}_i^{t_i^{end}}(1)) -  \frac{1}{N}\sum_{j=1}^N (Y_j^{t_j^k} - \widehat{Y}_j^{t_j^k}) 
    \end{align} 
    with sample variances \begin{align*}
        \widehat{\sigma}_{ATT}^2 &=\frac{1}{n-n_L-1}\sum_{i=n_L+1}^{n}({\Delta_{ATT}}_i - {\bar{\Delta}_{ATT}})^2 \\
        \widehat{\sigma}_{post}^2 &=\frac{1}{n_L-1}\sum_{i=1}^{n_L}({\Delta_{post}}_i - {\bar{\Delta}_{post}})^2 \\
        \widehat{\sigma}_{control}^2 &=\frac{1}{N-1}\sum_{j=1}^N({\Delta_{control}}_j - {\bar{\Delta}_{control}})^2
    \end{align*} 
    for ${\Delta_{ATT}}_i =\widehat{Y}_i^{t_i^{end}}(1) - \widehat{Y}_i^{t_i^{end}}(0)$ with mean $\bar{\Delta}_{ATT}$, ${\Delta_{post}}_i ={Y}_i^{t_i^{end}}(1) - \widehat{Y}_i^{t_i^{end}}(1)$ with mean $\bar{\Delta}_{post}$, and ${\Delta_{control}}_j ={Y}_j^{t_j^{k}} - \widehat{Y}_j^{t_j^{k}}$ with mean $\bar{\Delta}_{control}$, respectively. 
    
    Then with $z_{1-\alpha/2}$ denoting the $1-\alpha/2$ quantile of the standard normal distribution, the prediction powered confidence interval created \begin{align}
        \widehat{ATT}^{PPI} \pm z_{1-\alpha/2} \sqrt{\frac{\widehat{\sigma}_{ATT}^2}{n-n_L} + \frac{\widehat{\sigma}_{post}^2}{n_L} + \frac{\widehat{\sigma}_{control}^2}{N}}
    \end{align} has asymptotic coverage of ATT at level $1-\alpha$ assuming finite true variance and consistent variance estimators. \\

\textit{Proof.}
    Our estimator has three terms. We have $\widehat{ATT}^{PPI} = \Delta_{ATT} + \Delta_{post} + \Delta_{control}$, where each term is a sample mean over an iid sample of size $n-n_L$, $n_{L}$, or $N$ respectively. 
    
    We first show unbiasedness, i.e. $\mathbb{E}[\widehat{ATT}^{PPI}] = {ATT}$. Applying expectations throughout, we have
    \begin{align}
        \mathbb{E}[\widehat{ATT}^{PPI}] &= \mathbb{E}_{\mathcal{D}_{obs}}\left[\frac{1}{n-n_L}\sum_{i=n_L+1}^{n} (\widehat{Y}_i^{t^{end}_i}(1) - \widehat{Y}_i^{t^{end}}(0))\right] \\ &+  \mathbb{E}_{\mathcal{D}_{obs}}\left[\frac{1}{n_L}\sum_{i=1}^{n_L} (Y^{t^{end}_i}_i(1) - \widehat{Y}_i^{t^{end}_i}(1))\right] \\
        &- \mathbb{E}_{\mathcal{D}_{train}}\left[\frac{1}{N}\sum_{j=1}^N (Y_j^{t_j^k} - \widehat{Y}_j^{t_j^k}) \right].
            \end{align}
Since the post predictions $\widehat{Y}^{t^{end}}(1)$ are sampled independently from the same distribution $\mathcal{D}_{obs}$, we can cancel:
   
   \begin{align} &=\mathbb{E}_{\mathcal{D}_{obs}}\left[\frac{1}{n-n_L}\sum_{i=n_L+1}^{n} (- \widehat{Y}_i^{t^{end}_i}(0))\right] \\ 
        &+  \mathbb{E}_{\mathcal{D}_{obs}}\left[\frac{1}{n_L}\sum_{i=1}^{n_L} (Y^{t^{end}_i}_i(1))\right] 
        \\&- \mathbb{E}_{\mathcal{D}_{train}}\left[\frac{1}{N}\sum_{j=1}^N (Y_j^{t_j^k} - \widehat{Y}_j^{t_j^k}) \right]\\
        \end{align}
        
\noindent {With linearity of expectation and iid terms, we can cancel sums:}
        \begin{align}
        &= \mathbb{E}_{\mathcal{D}_{obs}}\left[Y^{t^{end}_i}_i(1)\right] - \mathbb{E}_{\mathcal{D}_{obs}}\left[\widehat{Y}_i^{t^{end}_i}(0)\right] \\
        \\&- \mathbb{E}_{\mathcal{D}_{train}}\left[(Y_j^{t_j^k} - \widehat{Y}_j^{t_j^k}) \right]
        \end{align}
\noindent {Using that the pre-treatment data are assumed to come from the same distribution as the training data:} \begin{align}
         &= \mathbb{E}_{\mathcal{D}_{obs}}\left[Y^{t^{end}_i}_i(1)\right] - \mathbb{E}_{\mathcal{D}_{obs}}\left[\widehat{Y}_i^{t^{end}_i}(0)\right] \\
        \\&- \mathbb{E}_{\mathcal{D}_{obs}}\left[(Y_j^{t_j^k} - \widehat{Y}_j^{t_j^k}) \right]
        \end{align}
    
\noindent {Since the $Y_i(0)$ in $\mathcal{D}_{obs}$ and the $Y_j$ in $\mathcal{D}_{train}$ come from the same distribution, we can then assuming matching time gaps and relabeling rewrite as follows:} 
         \begin{align}
         &= \mathbb{E}_{\mathcal{D}_{obs}}\left[Y^{t^{end}}_i(1)\right] - \mathbb{E}_{\mathcal{D}_{obs}}\left[\widehat{Y}_i^{t^{end}}(0))\right] \\
        \\&- \mathbb{E}_{\mathcal{D}_{obs}}\left[Y_i^{t_i^{end}}(0) - \widehat{Y}_i^{t_i^{end}}(0) \right]\end{align}

       \noindent  {And finally, canceling:} \begin{align}
        \\&= \mathbb{E}_{\mathcal{D}_{obs}}\left[Y^{t_i^{end}}_i(1)\right] -\mathbb{E}_{\mathcal{D}_{obs}}\left[Y_i^{t_i^{end}}(0) \right]\\
        &\text{which is the desired ATT.} \notag
    \end{align}

   The rest of the proof is standard arguments analogous to \cite{ppi}. The only difference is we combine three independent terms rather than two. For each sample's term above denoted as $\Delta$, assuming independence, iid terms, and finite variance, the Central Limit Theorem gives convergence for $n-n_L, n_L, N \to \infty$:
\begin{align}
    \sqrt{n-n_L}\,(\Delta_{ATT} - \mathbb{E}[\Delta_{ATT}]) &\Rightarrow \mathcal{N}(0, \sigma^2_{ATT}), \\
    \sqrt{n_{L}}\,(\Delta_{post} - \mathbb{E}[\Delta_{post}]) &\Rightarrow \mathcal{N}(0, \sigma^2_{post}), \\
    \sqrt{N}\,(\Delta_{control} - \mathbb{E}[\Delta_{control}]) &\Rightarrow \mathcal{N}(0, \sigma^2_{control}).
\end{align}

To combine the terms to a single interval, first, assume $n_L/(n-n_L) \to \rho_1$ and $N/(n-n_L) \to \rho_2$ for fixed constants $\rho_1, \rho_2 \in (0,1]$ as $n-n_L, n_{L}, N \to \infty$ for larger unlabeled data sample of size $(n-n_L)$. Rescaling each term to a common rate $\sqrt{n-n_L}$ and applying Slutsky's theorem,
\begin{align}
    \sqrt{n-n_L}\,(\Delta_{post} - \mathbb{E}[\Delta_{post}]) &\Rightarrow \mathcal{N}(0, \sigma^2_{post}/\rho_1), \\
    \sqrt{n-n_L}\,(\Delta_{control} - \mathbb{E}[\Delta_{control}]) &\Rightarrow \mathcal{N}(0, \sigma^2_{control}/\rho_2).
\end{align}
Since the three samples are mutually independent, the three rescaled sequences
converge jointly to independent normals and by the continuous
mapping theorem, their sum converges:
\begin{align}
    \sqrt{n-n_L}\big(\widehat{ATT}^{PPI} - \mathbb{E}[\widehat{ATT}^{PPI}]\big)
    \Rightarrow \mathcal{N}\!\left(0, \, {\sigma^2_{ATT}} + \frac{\sigma^2_{post}}{\rho_1} + \frac{\sigma^2_{control}}{\rho_2}\right).
\end{align}

Denote by $\widehat{\sigma}^2_{ATT}, \widehat{\sigma}^2_{post},
\widehat{\sigma}^2_{control}$ the corresponding sample variances. Then by assuming consistency of the sample variances for the true variance, we have, rescaling by $\frac{1}{n-n_L}$, that the interval
\begin{align}
    \widehat{ATT}^{PPI} \pm z_{1-\alpha/2} \sqrt{\frac{\widehat{\sigma}_{ATT}^2}{n-n_L} + \frac{\widehat{\sigma}_{post}^2}{n_L} + \frac{\widehat{\sigma}_{control}^2}{N}}
\end{align}
has asymptotic coverage of ATT at level $1-\alpha$. $\square$

\subsection{Covariate shift estimators}\label{cov_inf_proof}

We prove the result for the covariate shift estimator when we have some labels for the post treatment data, i.e. without assuming surrogacy. The result follows by analogous proof in the case where we adjust the training data on both ends (\textbf{Proposition 5}). The only difference is that in \textbf{Proposition 5} we must split the data distributed like $\mathcal{D}_{train}$ into two groups to maintain independent samples (i.e. of size $N/2$ and $N/2$). Again, the arguments used are standard and ultimately owe to \cite{ppi} and its extensions. \\

\noindent \textbf{Proposition 8.} \textit{(Covariate Shift Estimator without Surrogacy)} As stated above. \\

\textit{Proof.}
    Our estimator has three terms. Write $\widetilde{ATT}^{PPI-L} = \Delta_{ATT} + \Delta_{post} + \Delta_{control}$, where each term is a sample mean over an iid sample of size $n-n_L$, $n_L$, and $N$ respectively corresponding to the terms in Equation \ref{ATT_tilde}. 
    
    We first show $\mathbb{E}[\widetilde{ATT}^{PPI-L}] = {ATT} + o_p(N^{-1/2})$. Applying expectations throughout, we have
    \begin{align}
        \mathbb{E}[\widetilde{ATT}^{PPI-L}] &= \mathbb{E}_{\mathcal{D}_{obs}}\left[\frac{1}{n-n_L}\sum_{i=n_L+1}^{n} (\widehat{Y}_i^{t_i^{end}}(1) - \widehat{Y}_i^{t_i^{end}}(0))\right] \\ &+  \mathbb{E}_{\mathcal{D}_{obs}}\left[\frac{1}{n_L}\sum_{i=1}^{n_L} (Y^{t_i^{end}}_i(1) - \widehat{Y}_i^{t_i^{end}}(1))\right] \\
        &- \mathbb{E}_{\mathcal{D}_{train}}\left[\frac{1}{N}\sum_{j=1}^N \hat{w}(X_j^{t^0})\cdot (Y_j^{t_j^k} - \widehat{Y}_j^{t_j^k}) \right].
            \end{align}
Since the post predictions $\widehat{Y}^{t^{end}}(1)$ come from the same distribution $\mathcal{D}_{obs}$, we can cancel:
   
   \begin{align} &=\mathbb{E}_{\mathcal{D}_{obs}}\left[\frac{1}{n-n_L}\sum_{i=n_L+1}^{n} (- \widehat{Y}_i^{t_i^{end}}(0))\right] \\ 
        &+  \mathbb{E}_{\mathcal{D}_{obs}}\left[\frac{1}{n_L}\sum_{i=1}^{n_L} (Y^{t_i^{end}}_i(1))\right] 
        \\&- \mathbb{E}_{\mathcal{D}_{train}}\left[\frac{1}{N}\sum_{j=1}^N \hat{w}(X_j^{t^0})\cdot (Y_ij^{t_j^k} - \widehat{Y}_j^{t_i^k}) \right]\\
        \end{align}
        
\noindent {With linearity of expectation over iid terms, we can write:} \begin{align}
        &= \mathbb{E}_{\mathcal{D}_{obs}}\left[Y^{t_i^{end}}_i(1)\right] - \mathbb{E}_{\mathcal{D}_{obs}}\left[\widehat{Y}_i^{t_i^{end}}(0)\right] \\
        \\&- \mathbb{E}_{\mathcal{D}_{train}}\left[\hat{w}(X_j^{t^0})\cdot (Y_j^{t_j^k} - \widehat{Y}_j^{t_j^k}) \right]\\
        \end{align}
\noindent {We add and subtract the true density ratio term:}  \begin{align}
         &= \mathbb{E}_{\mathcal{D}_{obs}}\left[Y^{t_i^{end}}_i(1)\right] - \mathbb{E}_{\mathcal{D}_{obs}}\left[\widehat{Y}_i^{t_i^{end}}(0)\right] 
         \\&- \mathbb{E}_{\mathcal{D}_{train}}\left[\hat{w}(X_j^{t^0})\cdot (Y_j^{t_j^k} - \widehat{Y}_j^{t_j^k}) \right]
         \\&+ \mathbb{E}_{\mathcal{D}_{train}}\left[w(X_j^{t^0})\cdot(Y_j^{t_j^k} - \widehat{Y}_j^{t_j^k}) \right] \\&- \mathbb{E}_{\mathcal{D}_{train}}\left[w(X_j^{t^0})\cdot(Y_j^{t_j^k} - \widehat{Y}_j^{t_j^k}) \right]
         \end{align}
   \noindent      {Assuming comparability for pre-treatment data, we have $P(Y^{t^k}|X^{t^0})$ is constant across $\mathcal{D}_{obs}, \mathcal{D}_{train}$ hence we can change measure:} \begin{align}
         &= \mathbb{E}_{\mathcal{D}_{obs}}\left[Y^{t_i^{end}}_i(1)\right] - \mathbb{E}_{\mathcal{D}_{obs}}\left[\widehat{Y}_i^{t_i^{end}}(0))\right] \\
        \\&- \mathbb{E}_{\mathcal{D}_{obs}}\left[(Y_i^{t_i^k} - \widehat{Y}_i^{t_i^k}) \right]\\
        \\&+ \mathbb{E}_{\mathcal{D}_{train}}\left[(w(X_j^{t^0}) - \hat{w}(X_j^{t^0}))\cdot(Y_j^{t_j^k} - \widehat{Y}_j^{t_j^k}) \right]\end{align}
  \noindent  {Assuming matching time gaps and relabeling the error:} \begin{align}
         &= \mathbb{E}_{\mathcal{D}_{obs}}\left[Y^{t_i^{end}}_i(1)\right] - \mathbb{E}_{\mathcal{D}_{obs}}\left[\widehat{Y}_i^{t_i^{end}}(0)\right] 
        \\&- \mathbb{E}_{\mathcal{D}_{obs}}\left[Y_i^{t_i^{end}}(0) - \widehat{Y}_i^{t_i^{end}}(0) \right]
        \\&+ \mathbb{E}_{\mathcal{D}_{train}}\left[(w(X_j^{t^0}) - \hat{w}(X_j^{t^0}))\cdot e \right]\end{align}
{for error term $e=Y_i^{t^k_i} - \hat{Y}_i^{t^k_i}$ with $\mathbb{E}[e^2] < \infty$ so that $||e||_{L_2}=O_p(1)$\footnote{This is automatic if the outcomes and predictions are bounded.}. 
Then, canceling the expectation of the predictions $\hat{Y}^{t^{end}}(0)$:} \begin{align}
        \\&= \mathbb{E}_{\mathcal{D}_{obs}}\left[Y^{t_i^{end}}_i(1)\right] -\mathbb{E}_{\mathcal{D}_{obs}}\left[Y_i^{t_i^{end}}(0) \right] + \mathbb{E}_{\mathcal{D}_{train}}\left[(w(X_j^{t^0}) - \hat{w}(X_j^{t^0}))\cdot e \right]
         \end{align}

\noindent Notice that by our asymptotic assumptions, we have that the magnitude of the bias can be written
\begin{align}
   |\mathbb{E}[(w(X_j^{t^0}) - \hat{w}(X_j^{t^0})) \cdot e]| &  \leq \mathbb{E}[|(w(X_j^{t^0}) - \hat{w}(X_j^{t^0})) \cdot e|] \\
   &  \leq ||w(X_j^{t^0}) - \hat{w}(X_j^{t^0}) ||_{L_2} \cdot ||e||_{L_2}\\ % norm is an expectation 
   & = o_p(N^{-1/2}) \cdot O_p(1) = o_p(N^{-1/2})
\end{align}\footnote{We can relax this and obtain $\sqrt{N}$-consistency. Logistic regression (correctly specified) is, for example, $O_p(N^{-1/2})$, using this condition instead on $\hat{w}$, we get $\sqrt{N}$-consistency.}
{where the second inequality is by Cauchy-Schwartz. Then, we have $\mathbb{E}[\widetilde{ATT}^{PPI-L}] = {ATT} + o_p(N^{-1/2})$.} 

   For each sample's term above denoted as $\Delta$, assuming independence including of the data used to train $\hat{w}$, iid terms, and finite variance, the Central Limit Theorem gives convergence as $n-n_L, n_L, N \to \infty$:
\begin{align}
    \sqrt{n-n_L}\,(\Delta_{ATT} - \mathbb{E}[\Delta_{ATT}]) &\Rightarrow \mathcal{N}(0, \sigma^2_{ATT}), \\
    \sqrt{n_{L}}\,(\Delta_{post} - \mathbb{E}[\Delta_{post}]) &\Rightarrow \mathcal{N}(0, \sigma^2_{post}), \\
    \sqrt{N}\,(\Delta_{control} - \mathbb{E}[\Delta_{control}]) &\Rightarrow \mathcal{N}(0, \sigma^2_{control}).
\end{align} 

Assume $n_L/(n-n_L) \to \rho_1$ and $N/(n-n_L) \to \rho_2$ for fixed constants $\rho_1, \rho_2 \in (0,1]$ as $n, n_{L}, N \to \infty$ for large unlabeled data sample $n-n_L$. Rescaling each term to a common rate $\sqrt{(n-n_L)}$ and applying Slutsky's theorem,
\begin{align}
    \sqrt{(n-n_L)}\,(\Delta_{post} - \mathbb{E}[\Delta_{post}]) &\Rightarrow \mathcal{N}(0, \sigma^2_{post}/\rho_1), \\
    \sqrt{(n-n_L)}\,(\Delta_{control} - \mathbb{E}[\Delta_{control}]) &\Rightarrow \mathcal{N}(0, \sigma^2_{control}/\rho_2).
\end{align}
Since the three samples are mutually independent and the error is $o_p((n-n_L)^{-1/2})$, the three rescaled sequences
converge jointly to independent normals and by the continuous
mapping theorem their sum converges:
\begin{align}
    \sqrt{(n-n_L)}\big(\widetilde{ATT}^{PPI-L} - \mathbb{E}[\widetilde{ATT}^{PPI-L}]\big)
    \Rightarrow \mathcal{N}\!\left(0,\ {\sigma^2_{ATT}} + \frac{\sigma^2_{post}}{\rho_1} + \frac{\sigma^2_{control}}{\rho_2}\right).
\end{align}

The rest follows the same as the previous result. Denote by $\widehat{\sigma}^2_{ATT}, \widehat{\sigma}^2_{post},
\widehat{\sigma}^2_{control}$ the corresponding sample variances. Then by consistency of the sample variances and their combination, we have, rescaling by $\frac{1}{(n-n_L)}$ that the interval
\begin{align}
    \widetilde{ATT}^{PPI-L} \pm z_{1-\alpha/2} \sqrt{\frac{\widehat{\sigma}_{ATT}^2}{n-n_L} + \frac{\widehat{\sigma}_{post}^2}{n_L} + \frac{\widehat{\sigma}_{control}^2}{N}}
\end{align}
has asymptotic coverage of ATT at level $1-\alpha$. $\square$

\section{Two-Stage Estimation Approach for Heterogeneous Effects}\label{app:het_effects}

Our results also suggest a two stage modeling approach for learning heterogeneous treatment effects for prediction on new data without requiring labels in the treated population. In the case that the assumptions for causal inference with before and after AI surrogates are satisfied and $\widehat{\mu}$ is an excellent estimator of $\mu$, we can do the following. First, recall, we have access to an intervention-free training dataset $$\mathcal{D}_{train} = \{(X_i^{t_i^1},X_i^{t^2_i},\ldots, Y^{t^1_i}_i,Y^{t^2_i}_i, \ldots)\}_{i=1}^N,$$ and an (unlabeled) observational dataset $$\mathcal{D}_{obs}=\{(X^{t^{pre}}_i, X^{t^{post}}_i, A^{t_i}_i)\}_{i=1}^n.$$ We use the binary random variable $D$ to indicate the which dataset, with $D=1$ being the observational set $\mathcal{D}_{obs}$. The training dataset gives us knowledge about the processes $X^t, Y^t$ over time. The observed data has $X^t$ at two time points ${t^{pre}}, {t^{post}}$ relative to a binary treatment $A^t$ occuring by $t^{post}$.\\

\textbf{Stage 1: Surrogate Network}. This stage is the same as the model used in the main text. We train a longitudinal model $\widehat{\mu}(x,t)$ to predict $Y^{t'}$ over time $t'$ from a cross sectional data point $X^t$ using $\mathcal{D}_{train}$. In particular, we learn the model we can write as
\begin{align}
    \widehat{\mu}(x, t') \approx \mathbb{E}[Y^{t+t'}|X=x^t, D=0]
\end{align} which predicts $Y$ at time $t'$ years in the future from the observed $X$ for a series of times $t'$ through empirical risk minimization (ERM), i.e. 
\begin{align}
    \min_{\widehat{\mu}} \frac{1}{{N}}\sum_{i=1}^{{N}}\ell(\widehat{\mu}(x^{t_i}, t_i' - t_i), y^{t_i'})
\end{align} for some loss $\ell$. In the main text, for real world experiments, we learned the model using binary cross entropy for a binary outcome. For the simulations with continuous $Y^t$, we used mean squared error (MSE). \\

\textbf{Stage 2: CATT Network}. After training and fixing the surrogate network $\widehat{\mu}$, we run the predictions on $\mathcal{D}_{obs}$ to create estimates of CATT for each individual, i.e.~we create a new dataset $${\mathcal{D}_{obs}'} = \{(\widehat{\mu}(X^{t^{post}}_i, {t^{end}} - {t^{post}}) - \widehat{\mu}(X^{t^{pre}}_i, {t^{end}} - {t^{pre}})), X^{t^{pre}}_i\}_{i=1}^{{n}}$$ for any $t^{end}$ on which the surrogate network is trained. This dataset is used to train a second stage model predicting CATT from baseline measurements. Here we learn the model 
\begin{align}
     \widehat{\theta}(x) \approx \mathbb{E}[
    \mu(X^{t^{post}}, {t^{end}} - {t^{post}}) - \mu(X^{t^{pre}}, {t^{end}} - {t^{pre}})|X^{t^{pre}}=x, D=1]
\end{align}
through empirical risk minimization (ERM) for loss $\ell$ (such as MSE for the continuous outcome):
\begin{align}
    \min_{\widehat{\theta}} \frac{1}{n}\sum_{i=1}^{n}\ell(\widehat{\theta}(x_i), \widehat{\mu}(X^{t^{post}}_i, {t^{end}} - {t^{post}}) - \widehat{\mu}(x_i, {t^{end}} - {t^{pre}}))).
\end{align} 

Using the CATT network $\widehat{\theta}$, we can predict and evaluate CATT using new data having trained on $\mathcal{D}_{obs}$. This framework allows for estimating CATT in new populations with a similar distribution of $X^{t^{pre}}$ and is suitable only when assumptions for surrogacy are satisfied. It is particularly challenging to evaluate this approach in practice as we cannot evaluate the accuracy of effect estimation in real data. We can only evaluate how well the CATT model $\hat{\theta}$ approximates the effect predicted by $\widehat{\mu}$. However, this approach, in the case of strong assumptions, can lead to a method of modeling heterogeneous treatment effects for new individuals without ever observing the true outcomes for anyone.

\section{Further Experiments and Details}\label{app:exps}

\subsection{Synthetic Experiments}

\textit{\textbf{Details.}} We use a gradient-boosted decision tree as the AI model for our simulations. Each simulation runs a tree with up to 300 iterations using learning rate 0.5 with up to 31 maximum nodes each constructed with at least 20 samples. 

\noindent \textit{\textbf{Confounding experiments.}} One of the benefits of before-and-after AI surrogates, including over the AI surrogates in \cite{athey2025surrogate} is the relaxation of the unconfoundedness assumption. We consider simulations demonstrating this here. Consider the same data generation process as in the main text, however, we now add a confounding variable $C \sim \mathcal{N}(0,1)$ to the treatment probability in the observed dataset $\mathcal{D}_{obs}$ and the outcome as follows. We sample the treatment using a binomial distribution with probability defined from 
\begin{align}
    \mbox{logit}_p = \log(p_{random} / (1-p_{random})) + \gamma_C * C\\
    p_{treated} = \exp(\mbox{logit}_p)
\end{align} with $p_{random}=0.5$ and $\gamma_C=0.5$. Then for the same $C_i$, we define the outcome impact through
\begin{align}
    {Z_2}^0_i = {Z_2}^0_i + \gamma_C * C_i.
\end{align} The treatment bump in $Z^{t^{post}}_1$ for the non-linear experiments in the results reported here is 0.5. The bump in the linear experiments is increased to 0.8 to better visualize the bias.

\begin{table}[h!]
\centering
\tiny
\setlength{\tabcolsep}{5pt}
\begin{tabular}{cc|cc|ccc|c|cc}
\toprule
& & \multicolumn{2}{c|}{Model Prediction} & \multicolumn{3}{c|}{PPI} & Naive & \multicolumn{2}{c}{DIM} \\
$\rho_{X1}$ & $\rho_{X2}$ & MAE & $R^2$ & Coverage & Bias & Interval & Bias & Bias & Interval \\
\midrule
\multicolumn{10}{l}{\textbf{Linear DGP}} \\
\addlinespace[2pt]
0.5 & 0.5 & 1.489 (0.003) & 0.186 (0.002) & 86.0\% & 0.182 (0.026) & 0.784 (0.007) & -0.538 (0.005) & 0.276 (0.032) & 0.833 (0.006) \\
0.5 & 1 & 1.067 (0.002) & 0.553 (0.002) & 88.0\% & 0.024 (0.018) & 0.503 (0.005) & -0.504 (0.004) & 0.322 (0.031) & 0.841 (0.006) \\
0.85 & 0.85 & 0.934 (0.002) & 0.656 (0.001) & 92.0\% & 0.085 (0.016) & 0.549 (0.004) & -0.178 (0.004) & 0.304 (0.030) & 0.844 (0.006) \\
0.85 & 1 & 0.659 (0.002) & 0.826 (0.001) & 96.0\% & 0.026 (0.014) & 0.384 (0.003) & -0.154 (0.003) & 0.266 (0.028) & 0.843 (0.008) \\
1 & 0.5 & 1.150 (0.003) & 0.490 (0.003) & 82.0\% & 0.167 (0.029) & 0.699 (0.005) & -0.102 (0.013) & 0.246 (0.036) & 0.858 (0.008) \\
1 & 0.85 & 0.677 (0.002) & 0.816 (0.001) & 96.0\% & 0.064 (0.014) & 0.439 (0.003) & -0.012 (0.007) & 0.245 (0.029) & 0.856 (0.008) \\
1 & 1 & 0.074 (0.000) & 0.996 (0.000) & 100.0\% & 0.004 (0.003) & 0.139 (0.002) & 0.001 (0.002) & 0.262 (0.029) & 0.853 (0.008) \\
\addlinespace[3pt]
\multicolumn{10}{l}{\textbf{Non Linear DGP}} \\
\addlinespace[2pt]
0.5 & 0.5 & 1.383 (0.003) & 0.193 (0.002) & 84.0\% & 0.186 (0.022) & 0.655 (0.003) & -0.387 (0.004) & 0.246 (0.028) & 0.707 (0.003) \\
0.5 & 1 & 1.035 (0.002) & 0.518 (0.002) & 90.0\% & 0.026 (0.016) & 0.443 (0.003) & -0.361 (0.004) & 0.281 (0.026) & 0.710 (0.004) \\
0.85 & 0.85 & 0.853 (0.002) & 0.654 (0.001) & 92.0\% & 0.070 (0.014) & 0.477 (0.003) & -0.161 (0.004) & 0.264 (0.024) & 0.711 (0.003) \\
0.85 & 1 & 0.605 (0.002) & 0.815 (0.001) & 98.0\% & 0.021 (0.012) & 0.353 (0.002) & -0.140 (0.003) & 0.234 (0.023) & 0.708 (0.004) \\
1 & 0.5 & 1.057 (0.002) & 0.502 (0.002) & 82.0\% & 0.147 (0.024) & 0.580 (0.002) & -0.064 (0.007) & 0.217 (0.030) & 0.718 (0.004) \\
1 & 0.85 & 0.594 (0.002) & 0.822 (0.001) & 92.0\% & 0.063 (0.013) & 0.366 (0.002) & -0.019 (0.004) & 0.233 (0.024) & 0.717 (0.004) \\
1 & 1 & 0.059 (0.000) & 0.998 (0.000) & 100.0\% & 0.002 (0.001) & 0.115 (0.000) & 0.010 (0.002) & 0.231 (0.025) & 0.711 (0.004) \\
\addlinespace[3pt]
\bottomrule
\end{tabular}
\caption{Coverage, bias, and confidence interval width by confounded data generation process across linear, non-linear and $\rho_{X_1}, \rho_{X_2}$ to vary surrogacy. We use 5\% of the data as labeled. Bias and interval width are reported as mean with standard error across simulation runs.}
\label{tab:app_confounder_coverage_bias}
\end{table}

\begin{table}[h!]
\centering
\tiny
\setlength{\tabcolsep}{4pt}
\begin{tabular}{lc|ccc|ccc|c}
\toprule
& & \multicolumn{3}{c|}{PPI} & \multicolumn{3}{c|}{DIM} & Naive \\
DGP & Labeled data & Coverage & Bias & Interval & Coverage & Bias & Interval & Bias \\
\midrule
Linear & 1\% & 96.0\% & -0.024 (0.041) & 1.115 (0.026) & 90.0\% & 0.035 (0.070) & 1.937 (0.040) & -0.499 (0.004) \\
 & 10\% & 94.0\% & 0.011 (0.013) & 0.357 (0.003) & 60.0\% & 0.256 (0.021) & 0.602 (0.004) & -0.507 (0.004) \\
 & 50\% & 94.0\% & -0.004 (0.007) & 0.185 (0.000) & 6.0\% & 0.245 (0.010) & 0.269 (0.001) & -0.505 (0.004) \\
 & 100\% &  & &  & 0.0\% & 0.256 (0.007) & 0.191 (0.000) & -0.503 (0.005) \\
\addlinespace[3pt]
\midrule
\addlinespace[3pt]
Non-linear & 1\% & 92.0\% & -0.028 (0.037) & 0.990 (0.015) & 92.0\% & 0.051 (0.055) & 1.623 (0.021) & -0.357 (0.003) \\
 & 10\% & 92.0\% & 0.017 (0.012) & 0.315 (0.001) & 54.0\% & 0.246 (0.019) & 0.503 (0.002) & -0.363 (0.004) \\
 & 50\% & 98.0\% & -0.001 (0.006) & 0.165 (0.000) & 4.0\% & 0.227 (0.008) & 0.225 (0.000) & -0.360 (0.004) \\
 & 100\% & &  &  & 0.0\% & 0.237 (0.005) & 0.159 (0.000) & -0.361 (0.004) \\
\bottomrule
\end{tabular}
\caption{Coverage, bias, and confidence interval width by labeled data fraction for a fixed bad surrogate $\rho_{X_1}=0.5, \rho_{X_2}=1.0$ and confounding. Bias and interval width are reported as mean with standard error across simulation runs. Naive bias uses 100\% of the data as AI surrogates.}
\label{tab:app_bad_surrogate_sim_confounded}
\end{table}

\begin{table}[h!]
\centering
\tiny
\setlength{\tabcolsep}{4pt}
\begin{tabular}{lc|ccc|ccc|c}
\toprule
& & \multicolumn{3}{c|}{PPI} & \multicolumn{3}{c|}{DIM} & Naive \\
DGP & Labeled data & Coverage & Bias & Interval & Coverage & Bias & Interval & Bias \\
\midrule
Linear & 1\% & 100.0\% & 0.000 (0.005) & 0.171 (0.008) & 90.0\% & 0.278 (0.054) & 1.816 (0.041) & 0.001 (0.003) \\
 & 10\% & 100.0\% & -0.000 (0.002) & 0.136 (0.001) & 58.0\% & 0.263 (0.027) & 0.603 (0.004) & 0.002 (0.002) \\
 & 50\% & 100.0\% & 0.003 (0.004) & 0.171 (0.001) & 0.0\% & 0.262 (0.009) & 0.269 (0.001) & -0.001 (0.002) \\
 & 100\% &  &  & & 0.0\% & 0.246 (0.006) & 0.190 (0.000) & -0.005 (0.002) \\
\addlinespace[3pt]
\midrule
\addlinespace[3pt]
Non-linear & 1\% & 100.0\% & -0.003 (0.003) & 0.128 (0.001) & 90.0\% & 0.253 (0.048) & 1.564 (0.021) & 0.008 (0.002) \\
 & 10\% & 100.0\% & -0.001 (0.001) & 0.115 (0.000) & 52.0\% & 0.248 (0.022) & 0.503 (0.002) & 0.012 (0.001) \\
 & 50\% & 100.0\% & 0.000 (0.004) & 0.152 (0.000) & 0.0\% & 0.239 (0.008) & 0.225 (0.000) & 0.012 (0.002) \\
 & 100\% &  &  &  & 0.0\% & 0.225 (0.005) & 0.159 (0.000) & 0.010 (0.002) \\
\bottomrule
\end{tabular}
\caption{Coverage, bias, and confidence interval width by labeled data fraction for a fixed great surrogate $\rho_{X_1}=1.0, \rho_{X_2}=1.0$ and confounding. Bias and interval width are reported as mean with standard error across simulation runs. Naive bias uses 100\% of the data as AI surrogates.}
\label{tab:app_good_surrogate_sim_confounded}
\end{table}

\subsection{Cardio-Oncology Experiments} 

\textbf{\textit{Labels}}. We create diagnostic labels using ICD10 codes and MUSE labels with physician verification. Any records coded in ICD9 were converted to ICD10 then mapped in ICD10. The ICD mapping for heart failure included ICD I11.0, I13.0, I13.2, I50.0-50.9.

\noindent \textbf{\textit{Positivity checks.}} To visualize the distribution of ECGs across treated and untreated groups, we took the embedding from the end of the residual neural network's third layer. The output has dimension 80000 (128 by 625). We reduce the dimension of the embedding with principal component analysis (PCA) with 50 components. Then we run a t-SNE with two components and perplexity 50 over 1000 iterations. 

\noindent \textbf{\textit{Sequential modeling assumption checks}}. We test the sequential assumptions (Assumptions 3 and 4) using $\mathcal{D}_{train}$ by comparing the AI model $\widehat{\mu}$ to a second model $\widehat{\mu}'$ trained using pairs of ECGs to predict the same outcomes over time. The second model uses the same residual neural network head for each ECG. Embeddings after the residual trunk are combined prior to the model disease heads. We predict from the ECG and its closest preceding ECG for all patients with multiple ECGs. If the previous image adds no predictive value, then Assumption 3 is satisfied. Indeed we see that the predictive performance by AUROC of the two models are very similar averaged across years. The average absolute difference across 15 years was 0.016 (0.005). The same auxiliary model evaluated instead for the predictive aide that the later ECG provides shows that Assumption 4 holds in the intervention-free data. Using the time window between the pairs of two years, we compare the AUROC for the auxiliary model shifted to be evaluating as if from the time of the earlier ECG to the primary model and again notice minimal change in predictability. The average absolute difference across 15 years was 0.013 (0.006).

\begin{figure}[h!]
    \centering
    \includegraphics[width=0.6\linewidth]{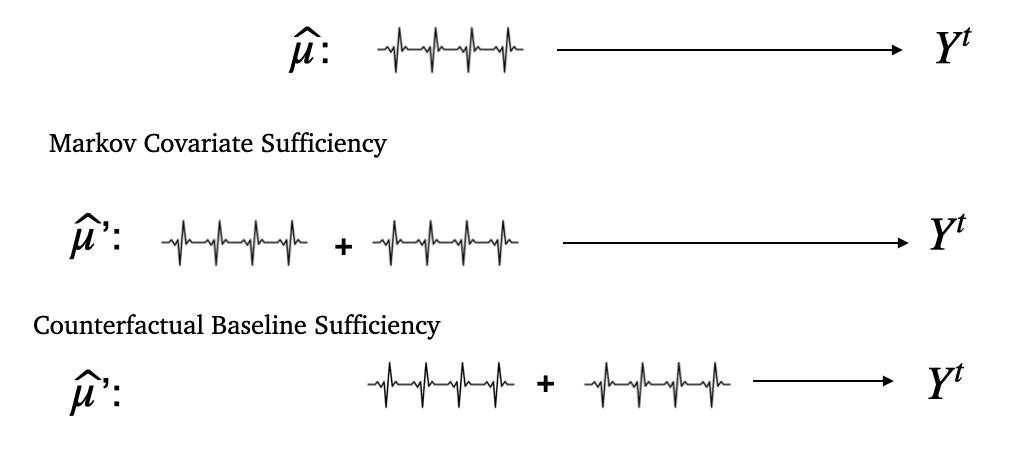}
    \caption{\textbf{Check for sequential model assumptions}. We train separate models that predict the same outcome from ECGs now with the addition of an earlier or later ECG for patients with multiple ECGs in the training data $\mathcal{D}_{train}$. The difference in AUROC between models over years is small suggesting minimal gain in adding these data in predicting $Y^t$.}
    \label{fig:seq_assmptns}
\end{figure}

\begin{figure}[h!]
    \centering
    \includegraphics[width=\linewidth]{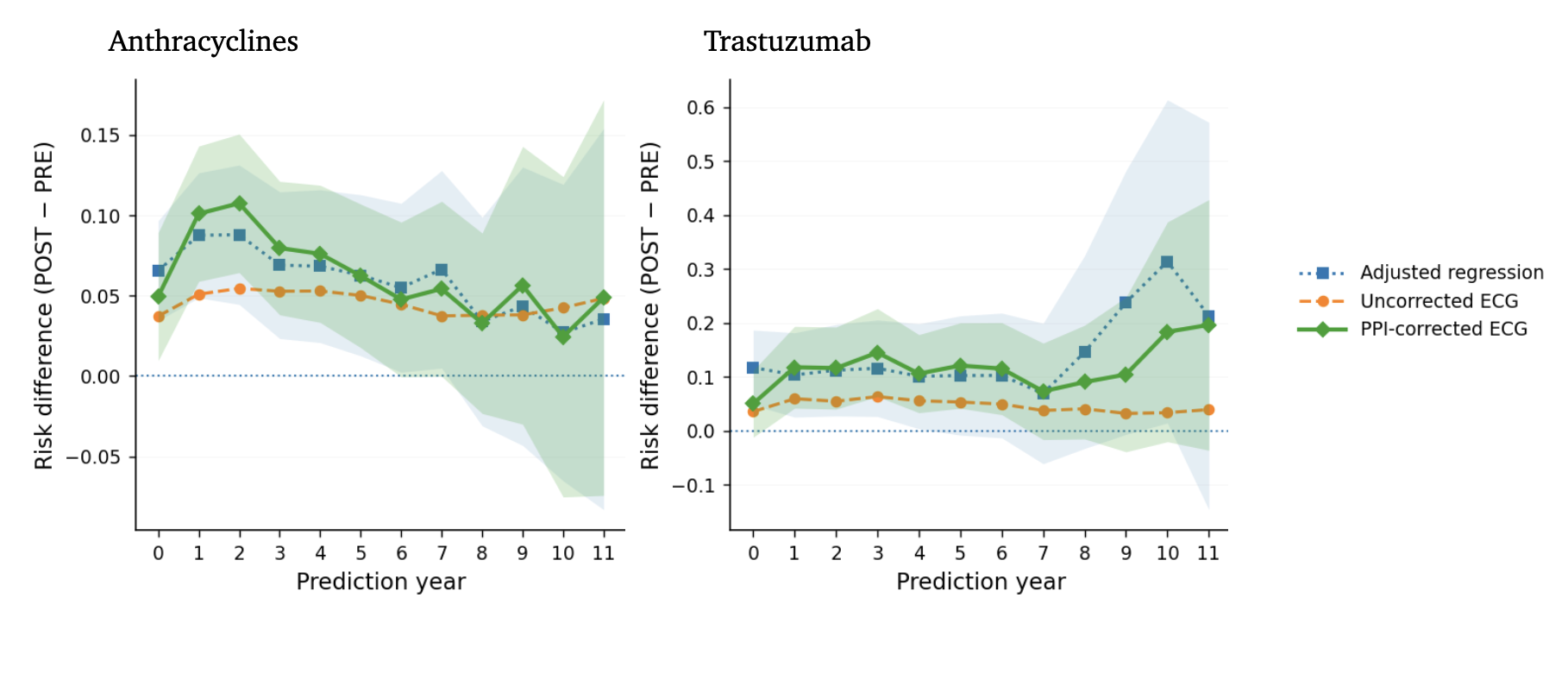}
    \caption{\textbf{Covariate shifted cardio-oncology results.} Debias correcting with the covariate shift for anthracycline and trastuzumab treatments gives nearly the same results as the data are not well distinguished from the test set.}
    \label{fig:no_cov_shift}
\end{figure}

\noindent \textbf{\textit{Covariate shift modeling}}. We explore covariate shift corrections for the real data. For the covariate shift models, which learn an estimate of the density ratio $w(x)$ between the pre and control (test set) ECGs, we train a logistic regression classifier using all pre treatment data and all other model test set data. Rather than giving the model the full layer 3 embedding, we use PCA again to reduce to 1000 components before inputting to the classifier. The classifier is trained using five fold cross validation. A classifier for the anthracycline and trastuzumab treated groups has AUROC of 0.557 and 0.517, respectively. The covariate shift therefore adds little in these cases and is omitted with nearly identical results. We show adding this covariate shift in Figure \ref{fig:no_cov_shift}. \\

\begin{figure}[h!]
    \centering\includegraphics[width=\linewidth]{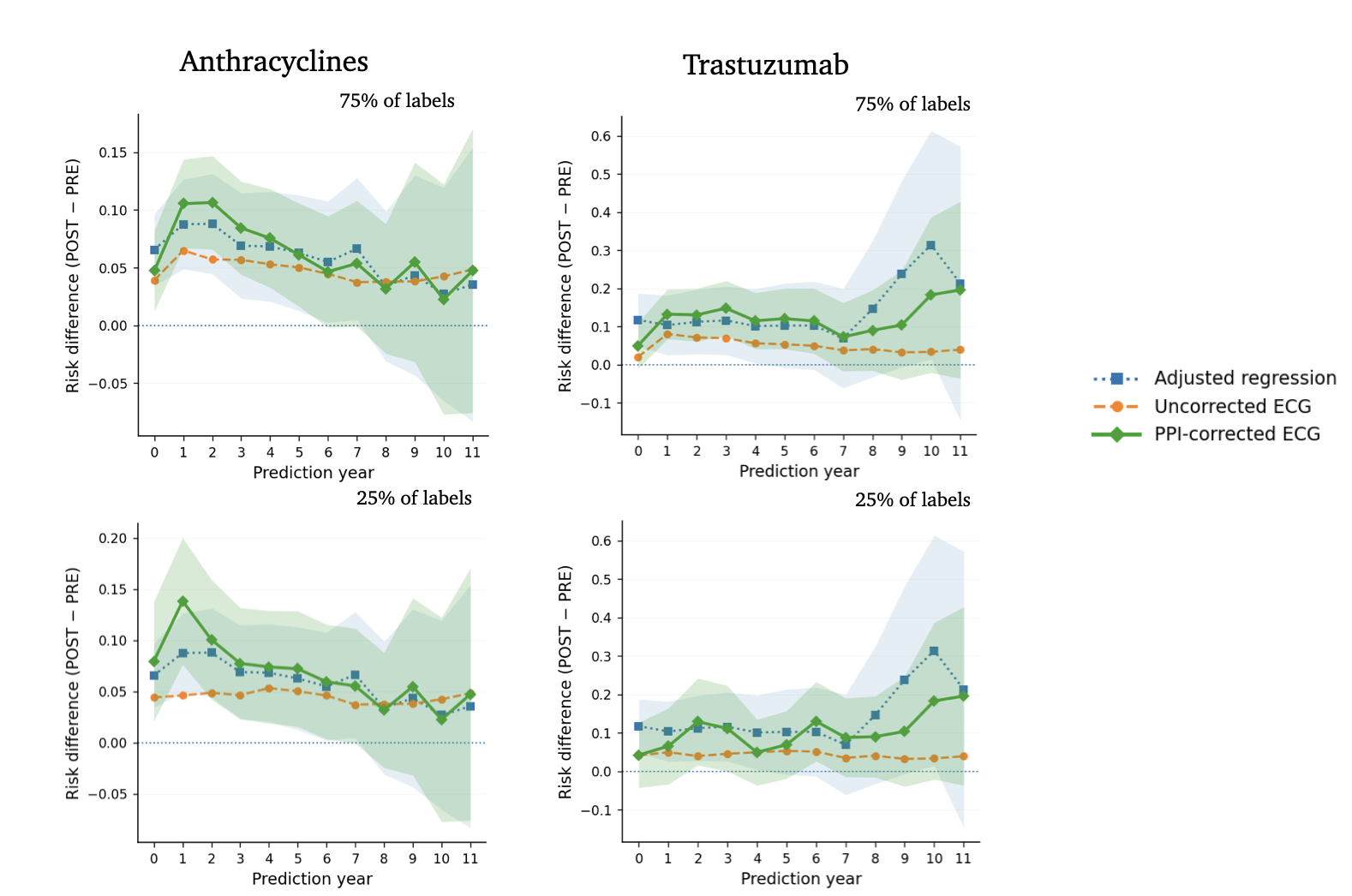}
    \caption{\textbf{Varying sample size of labeled treated individuals for cardio-oncology experiments}. These compare to the 50\% of labels used in the main text figure and have sample sizes as in Table \ref{app:tab:counts}.}
    \label{app:fig:counts}
\end{figure}

\noindent \textbf{\textit{Varying data quantity used for debias correction in cardio-oncology experiments}}. In the main text, we use 50\% of the patient's follow up labels to perform the debias correction. This gives fewer than 50\% of the total patients as having labels in the later years due to real loss to follow up (Table \ref{app:tab:counts}). We compare having more and fewer labels in Figure \ref{app:fig:counts}.

\begin{table}[h!]
    \centering
    \begin{tabular}{|c|c|c|c||c|c|c|c|}
        \hline
        \multicolumn{4}{|c||}{\textbf{Anthracycline treated}} &
        \multicolumn{4}{c|}{\textbf{Trastuzumab treated}} \\
        \hline
        \textit{Year} & \textbf{25\%} & \textbf{50\%} & \textbf{75\%} &
        \textit{Year} & \textbf{25\%} & \textbf{50\%} & \textbf{75\%} \\
        \hline
        0  & 161 & 322 & 483 & 0  & 48 & 97 & 146 \\
        1  & 161 & 322 & 448 & 1  & 48 & 97 & 146 \\
        2  & 161 & 322 & 396 & 2  & 48 & 97 & 138 \\
        3  & 161 & 322 & 348 & 3  & 48 & 97 & 129 \\
        4  & 161 & 297 & 297 & 4  & 48 & 97 & 99  \\
        5  & 161 & 251 & 251 & 5  & 48 & 79 & 79  \\
        6  & 161 & 213 & 213 & 6  & 48 & 71 & 71  \\
        7  & 161 & 174 & 174 & 7  & 48 & 56 & 56  \\
        8  & 127 & 127 & 127 & 8  & 43 & 43 & 43  \\
        9  & 67  & 67  & 67  & 9  & 28 & 28 & 28  \\
        10 & 40  & 40  & 40  & 10 & 19 & 19 & 19  \\
        11 & 22  & 22  & 22  & 11 & 12 & 12 & 12  \\
        \hline
    \end{tabular}
    \caption{\textbf{Sample sizes for debias correction's treated labels in cardio-oncology experiments}. We use 25 to 75\% of the data to explore the quantity of labels needed to debias the AI surrogate. We consider the percentage to be of all patients and then use real data masks from real follow up. This gives fewer samples in later years by design.}
    \label{app:tab:counts}
\end{table}

\end{document}